\documentclass{article}
\usepackage{lipsum}
\usepackage{amsmath,amssymb,amsfonts,amsthm}
\usepackage{graphicx}
\usepackage{epstopdf}
\usepackage{algorithmic}
\usepackage{algorithm}
\usepackage{color}
\usepackage{cleveref}
\usepackage{bm}
\ifpdf
  \DeclareGraphicsExtensions{.eps,.pdf,.png,.jpg}
\else
  \DeclareGraphicsExtensions{.eps}
\fi

\usepackage{enumitem}
\setlist[enumerate]{leftmargin=.5in}
\setlist[itemize]{leftmargin=.5in}

\usepackage{geometry}
\DeclareMathOperator*{\argmax}{argmax}

\newcommand{\RR}{\mathbb{R}}
\newcommand{\sign}{\text{sign}}
\newcommand{\E}{\mathbb{E}}
\newcommand{\Pm}{\mathbb{P}}

\newcommand{\thup}{^{\text{th}}}
\newcommand{\stup}{^{\text{st}}}
\newtheorem{theorem}{Theorem}
\def \sign {{\rm sign}}
\def \SCB {{SCB}}

\newcommand{\vct}[1]{\bm{#1}}
\newcommand{\mtx}[1]{\bm{#1}}

\def \A {\mtx{A}}

\def \X {\mtx{X}}
\def \M {\mtx{M}}
\def \Q {\mtx{Q}}

\def \y {\vct{y}}
\def \r {\vct{r}}
\def \x {\vct{x}}
\def \b {\vct{b}}
\def \q {\vct{q}}
\def \g {\vct{g}}

\Crefname{subsection}{Subsection}{Subsections}

\begin{document}

\title{An iterative method for classification of binary data}

 \author{Denali Molitor \quad and \quad Deanna Needell \\
 University of California, Los Angeles \\ Los Angeles, CA 90095, USA}

\date{}
\maketitle

\begin{abstract}
In today's data driven world, storing, processing, and gleaning insights from large-scale data are major challenges. Data compression is often required in order to store large amounts of high-dimensional data, and thus, efficient inference methods for analyzing compressed data are necessary. 
Building on a recently designed simple framework for classification using binary data, we demonstrate that one can improve classification accuracy of this approach through iterative applications whose output serves as input to the next application. As a side consequence, we show that the original framework can be used as a data preprocessing step to improve the performance of other methods, such as support vector machines. For several simple settings, we showcase the ability to obtain theoretical guarantees for the accuracy of the iterative classification method. The simplicity of the underlying classification framework makes it amenable to theoretical analysis and studying this approach will hopefully serve as a step toward developing theory for more sophisticated deep learning technologies. 

\end{abstract}

\section{Introduction}
We consider the problem of performing classification when only binary measurements of data are available. This situation may arise due to the need for extreme compression of data or in the interest of hardware efficiency \cite{fang2014sparse,JacquLBB_Robust,LaskaWYB_Trust,aziz1996overview}. Despite this extremely coarse quantization of the data, one can still perform learning tasks, such as classification, with high accuracy.
The authors of \cite{NSW17Simple} recently proposed a classification method for binary data, which they show to be reasonably accurate and sufficiently simple to allow for theoretical analysis in certain settings. Additionally, the predicted class can be approximately understood as the class whose binarized training data 
most closely and frequently matches that of the test point.   As this approach will be the foundation of the work presented here, we discuss it in detail in the next section.

Interpretability of algorithms and the ability to explain predictions is of increasing importance as machine learning algorithms are applied to an expanding range of problems in areas such as medicine, criminal justice, and finance \cite{barocas2016big,barocas2017big,executive2016big}. Decisions made based on algorithmic predictions can have profound repercussions for both participating individuals as well as society at large. A major drawback to complex models such as deep neural networks \cite{lecun1998gradient,he2016deep,collobert2008unified,lecun2015deep} is that it is extremely difficult to explain how or why such algorithms arrive at a specific prediction, see e.g. \cite{zhang2017interpretable,zhang2018visual,ventura2017interpreting} and references therein.  
Studying and advancing models for which model output can be understood will help to both improve methods that are more readily interpretable and develop tools for understanding more complex models. The aim of this paper is to continue developing a framework with these two simultaneous goals in mind.

\subsection{Contribution}
We propose an extension of the simple classification method for binary data proposed in \cite{NSW17Simple}, which we will henceforth refer to as \SCB.   
We find that our extension 
 often leads to improved performance over \SCB.  
 Additionally, we demonstrate that \SCB\ can be used for dimension reduction or as a data preprocessing step to improve the performance of other algorithms, such as support vector machines (SVM). 
  The proposed extension to \SCB\ that we consider here utilizes iterative applications of the original approach, reminiscent of the compositional nature of neural networks. Due to the simplicity of the \SCB\ framework, we can provide theoretical guarantees for the accuracy of the iterative extension in simple settings. We believe that studying this kind of iterative classification framework is interesting and practical in its own right, and will also serve as a step toward gaining a more thorough understanding of more complex deep learning strategies.

\subsection{Organization}
The paper is organized as follows. \Cref{sec:alg} introduces the problem statement and classification strategies of interest. \Cref{sec:orig_alg} describes the \SCB\ framework introduced in \cite{NSW17Simple} and \Cref{sec:iter_alg} our proposed iterative extension. In \Cref{sec:exper}, we demonstrate the performance of the proposed approach on real and synthetic datasets. \Cref{sec:rlg} discusses variations and practical considerations. We provide theoretical guarantees for the proposed iterative method in several simplified settings and provide intuition as to why the iterative method generally outperforms the original approach in \Cref{sec:theory}.  Finally, \Cref{sec:prepr} demonstrates how \SCB\ can be adapted to serve as a data preprocessing and dimension reduction strategy for other methods applied to binary data.

\section{Classification using binary data}
\label{sec:alg}

We first introduce the problem and notation that will be used throughout. Let $\A\in\RR^{m\times n}$ be a random measurement matrix (e.g. typically it will contain i.i.d. standard normal entries). Let $\X = [\x_1 \cdots \x_p]\in \RR^{n\times p}$ be the matrix of $p$ data vectors $\x_i\in \RR^n$ with labels $\b = (b_1,\cdots b_p)$. Let $G$ be the number of groups or classes to which the data points belong, so that we may assume $b_i\in\{1,2,\ldots,G\}$. 
Suppose we have the binary measurements of the data
\[\Q = \sign(\A\X),\] 
where $\sign(\M)_{i,j} = \sign(M_{i,j})$ and for a real number $c$ the $\sign$ function simply assigns $\sign(c)=1$ if $c\geq 0$ and $-1$ otherwise. For a matrix $\M$, 
let $\M_{(j)}$ denote the $j\thup$ column of $\M$. 

The rows of the matrix $\A$ can be viewed as the normal vectors to randomly oriented hyperplanes, in which case the $(i,j)\thup$ entry of $\Q$ denotes on which side of the $i\thup$ hyperplane the $j\thup$ data point $\x_j$ lies. In practice, the binary data $\Q$ may be obtained during processing or be provided as direct input from some other source.  In the latter case, we may not have access to the data matrix $\X$ or the measurement matrix $\A$, but only the resulting binary data $\Q$. We refer to the binary information indicating the position of a data point relative to a set of hyperplanes as a \textit{sign pattern}. In particular, for a column $\Q_{(j)}$ and any subset of its entries, the resulting vector indicates the sign pattern of the $j\thup$ data point relative to that subset of hyperplanes.

\begin{figure}
\centering
\includegraphics[width = 0.4\textwidth]{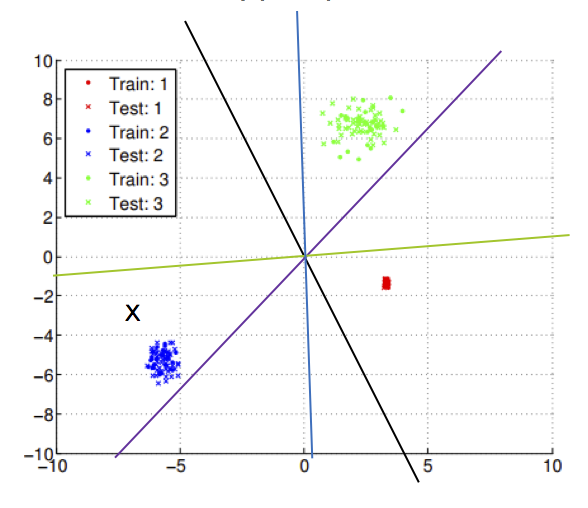}
\includegraphics[width = 0.4\textwidth]{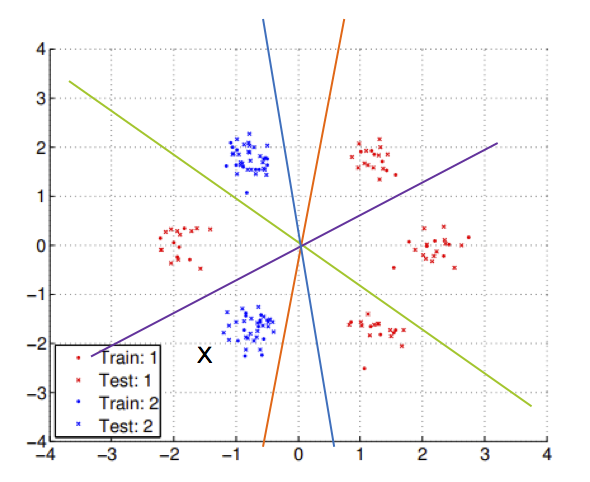}
\caption{A motivating example for using positions relative to hyperplanes for classification.}
\label{fig:motiv1}
\end{figure}

We aim to classify a data point $\x$ based only on the binary information contained in $\Q$. As a simple motivating example, consider the left plot of \Cref{fig:motiv1}. The training data points each belong to one of three classes, red, blue, or green. Consider the test point indicated by the black $\x$. Cycling through the hyperplanes, the green hyperplane indicates that the test point likely belongs to the blue or red class (since it lies on the same side as these clusters), the purple hyperplane indicates that the test point likely belongs to the blue or green class, the blue hyperplane indicates that the test point likely belongs to the blue class, and the black hyperplane indicates that the test point likely belongs to the blue class. In aggregate, the test point matches the relative positions of the blue class to the hyperplanes most often. This prediction matches what we might predict visually.

For the data in the right plot of \Cref{fig:motiv1}, there are both red and blue points on the same side as $\x$ for each hyperplane. However, if we consider sign patterns with respect to \textit{pairs} of hyperplanes instead of only single hyperplanes, we can isolate data within cones or wedges as opposed to simply half-spaces. Comparing the sign patterns of the training data with respect to pairs of hyperplanes with that of the test point, we find that the test point $\x$ matches the sign patterns of the blue class most often. Thus, it may not be enough to consider hyperplanes individually, but in tuples. 
\SCB\ uses this intuition as motivation. 

\subsection{Simple classification for binary data (SCB)}
\label{sec:orig_alg}

In SCB, sign patterns of the data with respect to tuples of hyperplanes of various lengths are recorded and aggregated to arrive at a prediction. The length of the sign patterns, or the number of hyperplanes considered, is referred to as the \textit{level}. For each level $\ell = 1,\cdots, L$,  we choose $m$ random combinations of $\ell$ hyperplanes. Each of the hyperplane-tuples provides a {\it measurement} of the data points. Fixing the number of hyperplane combinations considered, as opposed to considering all possible combinations, prevents the number of measurements from growing exponentially with the level.

Let $t$ be a sign pattern for the $i\thup$ measurement at the $\ell\thup$ level and $P_{g|t}$ be the number of training points in class $g$ with sign pattern $t$. 
This sign pattern information is then aggregated for the training data points in the membership function $\r(\ell,i,t,g)$, with 
\begin{equation}\label{eqn:RF}
\r(\ell,i,t,g) := \frac{P_{g|t}}{\sum_{j=1}^G P_{j|t}}\frac{\sum_{j=1}^G |P_{g|t}-P_{j|t}|}{\sum_{j=1}^G P_{j|t}},
\end{equation}
where $\ell = 1,\cdots, L$,  $i = 1,\cdots m$, and $g=1,\cdots, G$. The first term in this formula gives the fraction of points with sign pattern $t$ that belong to class $g$, while the second acts as a balancing term to account for differences in the relative sizes of different classes.
Each value in this membership function gives an indication of how likely a data point is to belong to class $g$ based on the fact that it has sign pattern $t$ for the $i\thup$ measurement at the $\ell\thup$ level. Larger $\r(\ell, i,t,g)$ values indicate that a data point is more likely to belong to the $g\thup$ class. 
Training is detailed in \Cref{algo:Train}, which simply computes all of these quantities.

Given a test point $\x$, with binary data $\q=\sign(\A\x)$, for each level $\ell$, measurement $i$ and associated sign pattern $t^*$ we find the corresponding $\r(\ell,i,t^*,g)$ value and keep a running sum for each group $g$, stored in the vector $\tilde \r$ (note that the vector $\tilde{\r}$ depends on the data point $\x$, but we notationally ignore this dependence for tidiness, and will write $\tilde{\r}(g)$ for a class $g$ or data point $\y$ when clarification is needed).
If $t^*$ does not match any of the sign patterns observed in the training data, then no update to $\tilde \r$ is made. The testing procedure is detailed in \Cref{algo:Class}. In
\cite{NSW17Simple}, the authors showed that this classification method works well on both artificial and real datasets (e.g. MNIST \cite{MNIST}, YaleB \cite{YaleB1,YaleB2,YaleB3,YaleB4}).

\begin{algorithm}[ht]
\begin{algorithmic}
    \caption{\SCB\ Training from  \cite{NSW17Simple}}
	\label{algo:Train}
    \STATE{\textbf{Input:} binary training data $\Q$, training labels $\b$, number of classes $G$, number of levels $L$.}
    \FOR{$\ell$ from 1 to $L$, $i$ from 1 to $m$}
       \STATE{Randomly select $\ell$ hyperplanes.}
	\FORALL{observed sign patterns $t$ and classes $g$ from $1$ to $G$}
		\STATE{Compute $\r(\ell,i,t,g)$ as in \Cref{eqn:RF}.}
	\ENDFOR
      \ENDFOR
\end{algorithmic}
\end{algorithm}

\begin{algorithm}[ht]
\begin{algorithmic}
    \caption{\SCB\ Classification from  \cite{NSW17Simple}}
	\label{algo:Class}
    \STATE{\textbf{Input:} binary testing data $\q$, number of classes $G$, number of levels $L$, learned parameters $\r(\ell,i,t,g)$, and hyperplane tuples from \Cref{algo:Train}.}
    \FOR{$\ell$ from 1 to $L$, $i$ from 1 to $m$}
       \STATE{Identify the sign pattern $t^*$ to which $\q$ corresponds for the $i\thup$ $\ell$-tuple of hyperplanes. }
	\FOR{$g$ from $1$ to $G$}
		\STATE{$\tilde \r(g)=\tilde \r(g)+\r(\ell,i,t^*,g)$.}
	\ENDFOR
	\ENDFOR
	\STATE{Set $\tilde \r(g) = \frac{\tilde \r(g)}{Lm}$ for $g=1,\cdots,G$.}
	\STATE{ Classify $\hat b=\argmax_{g\in\{1,\cdots, G\}} \tilde \r(g)$. } 
\end{algorithmic}
      
\end{algorithm}

\subsection{Iterative classification for binary data (ISCB)}
\label{sec:iter_alg}

\begin{figure}[ht]
\centering
\includegraphics[width = 0.5\textwidth]{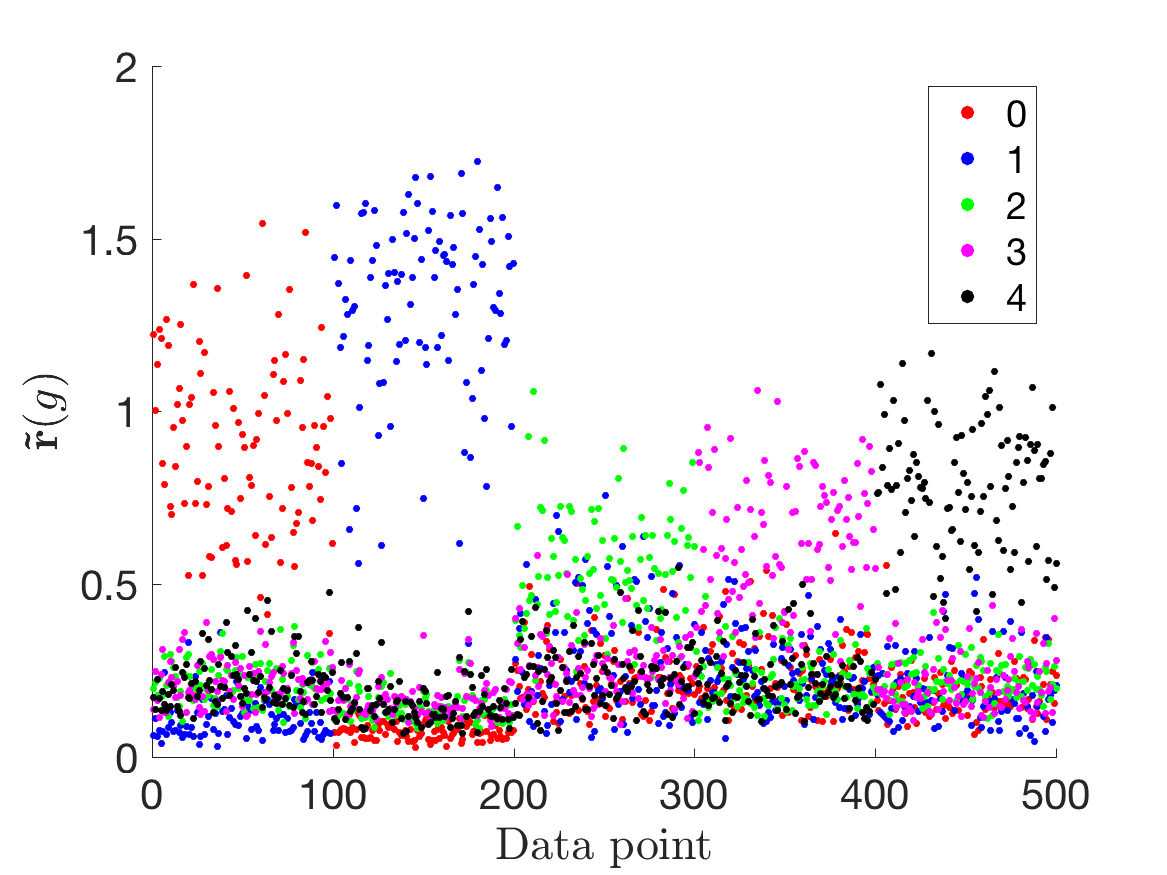}
\caption{The $\tilde \r(g)$ values from \SCB\ trained to classify digits 0-4 from the MNIST dataset are plotted. Five digits are considered to ease visualization. One-hundred test points from each digit are used with points 1-100 corresponding to 0s, 101-200 corresponding to 1s, etc. $\tilde \r(0)$ values are plotted in red, $\tilde \r(1)$ in blue, $\tilde \r(2)$ in green, $\tilde \r(3)$ in magenta and $\tilde \r(4)$ in black. }
\label{fig:rtilde}
\end{figure}

First, we motivate the iterative extension to \SCB\ through an example. Consider \Cref{fig:rtilde}, which plots the values of $\tilde \r$ from \Cref{algo:Class} for the task classifying the digits 0-4 of the MNIST dataset (where we will use class labels $0,1,\ldots,4$). Note that test images of the digit 0 typically have lower $\tilde \r(1)$ values than do other digits. Similarly, test images of the digit 1 typically have lower  $\tilde \r(0)$ than do test images of the digits 1-4. Indeed, it is not only likely that 
\[\hat b_{\x}=\argmax_{g\in\{1,\cdots, G\}} \tilde \r(g)\] 
corresponds to the true digit label, but in addition the $\tilde \r$ vectors for testing images from different digits contain different \textit{patterns}. Thus, we expect that using a method more advanced than simply predicting the class corresponding to the maximum of the the $\tilde \r$ vector may improve classification accuracy, specifically a strategy that makes use of the distribution of the values contained in $\tilde \r$.

One could make predictions based on the $\tilde \r$ vectors in a variety of ways. We mention a few such options here. Drawing intuition from simple neural network architectures such as multi-layer perceptron \cite{cybenko1989approximation} and boosting algorithms such as AdaBoost \cite{freund1997decision,freund1999short}, we first consider using iterative applications of \SCB, where $\tilde \r$ values of the training data from previous iterations are used as input training data for the following iteration. In particular, the method is reminiscent of the structure of a single neuron in a neural network in which information only propagates forward as opposed to throughout the whole network. In contrast to deep neural networks, the output at each iteration, $\tilde \r$, can be interpreted as a vector indicating to which class a data point $\x$ is likely to belong. This iterative strategy also relates to boosting in that subsequent iterations train on the shortcomings of previous iterations. Specifically, if points from a given class are misclassified, but produce similarly structured $\tilde \r$ vectors this pattern may be corrected in the next application of the algorithm.

The training and testing phases of the iterative version of \SCB, which we refer to as ISCB, are detailed in \Cref{algo:ItTrain} and \Cref{algo:ItClass}. 
To ease notation, let $\r_k$, $\tilde \r_k$, and $\A^{(k)}$ be $\r$, $\tilde \r$, and $\A$ from the $k\thup$ application of \SCB\  (\Cref{algo:Train} and \Cref{algo:Class}). During training, the first iteration in ISCB is executed as in \Cref{algo:Train}. We collect the data
$\X = [\tilde \r_1(\x_1) \cdots \tilde \r_1(\x_p)] \in \RR^{G\times p}$, which will be used as training data for the next iteration, where $\x_i$ are training data points. In contrast to \SCB, the iterative algorithm calculates $\tilde \r$ values for both the training and test data. Note that the dimension of the data points is fixed at $G$ after the first application of \SCB. For high dimensional data, we will typically have $G\ll n$. This reduction in dimension reduces the computational cost of some of the required computations, such as $\Q = \sign(\A\X).$ One could also make use of the same measurement matrix $\A$ for all iterations after the first. Since the dimension is much smaller after the first iteration of SCB, one may also need fewer levels for accurate classification.  We leave an exhaustive study of the many possible variations for future work, and focus here on establishing the mathematical framework of this iterative approach.

 After each application of \Cref{algo:Train}, we collect sign information of our data with respect to a new set of random hyperplanes. Although the dimension of the data for the subsequent applications lies in $\RR^G$ and thus we expect the size of this data to be manageable, there are several motivations for taking binary measurements of the data at each application. First, we can still take advantage of methods for efficient storage of and computation with binary data. Second, the binary measurements roughly preserve angular information about the data. For the $\tilde \r$ values, we are generally interested in the relative sizes of the components, since these represent the likelihood that a point belongs to a given class. The overall magnitude of the $\tilde \r$ values is of less importance and, thus, binary measurements retain the significant information pertaining to the data. Third, considering binary measurements of the data at each application maintains consistency between the applications, making the method more amenable to theoretical analysis, interpretability, and is more in line with sophisticated deep neural net architectures.

Since the components of $\tilde \r$ are always non-negative, we restrict the random hyperplanes to intersect this space after the first application.  
For example, we can ensure the hyperplanes intersect this region by requiring that the normal vectors have at least one positive and one negative coordinate. 
We do not recenter the data after each application, as the structure of $\tilde \r$ can lead to poor performance with recentering. For an example, consider \Cref{fig:dataprogr}, in which the $\tilde \r$ values follow a roughly linear trend for later applications of the method. 
Finally, after the last application of \SCB, for the iterative algorithm, we make the prediction \[ \hat b=\argmax_{g\in\{1,\cdots, G\}}  \tilde \r_K(g).\]

\begin{algorithm}[ht]
\begin{algorithmic}
    \STATE{\textbf{Input:} binary training data $\Q\in \RR^{m\times p}$, training labels $\b$, number of classes $G$, number of levels $L$, number of applications $K$.}
    \FOR{$k$ from 1 to $K$,}
      
       \STATE{Train learned parameters $\r_k(\ell,i,t,g)$ as in \Cref{algo:Train}, with input: $\Q$, $\b$, $G$ and $L$.}
	\STATE{Set $\X=0\in \RR^{G\times p}$.}
	\FOR{$j$ from 1 to $p$}
		\STATE{Apply \Cref{algo:Class} to $\Q_{(j)}$ using learned parameters $\r_k(\ell,i,t,g)$ to calculate $\tilde \r_k$.}
		\STATE{Set $\X_{(j)} = \tilde \r_k$.}
	\ENDFOR
	\STATE{Form the random measurement matrix $\A^{(k)}\in \RR^{m\times G}.$}
	\STATE{Set $\Q= \sign (\A^{(k)} \X)$.} 
      \ENDFOR
\STATE{\textbf{Output:} $\r_k(\ell,i,t,g)$, and $\A^{(k)}$ for $k$ from 1 to $K$. }
      \caption{ISCB Training. }
	\label{algo:ItTrain}
\end{algorithmic}
\end{algorithm}

\begin{algorithm}[ht]
\begin{algorithmic}
    \caption{ISCB Testing. } 
\label{algo:ItClass}
    \STATE{\textbf{Input:} binary test data $\q\in\RR^{m}$, number of classes $G$, levels $L$, and iterations $K$, learned parameters $\r_k(\ell,i,t,g)$, hyperplane tuples, and $\A^{(k)}$ from \Cref{algo:ItTrain}.}
	\FOR {$k$ from 1 to $K$}
	\STATE{Set $\tilde \r_k = 0$. } 
      \FOR{$\ell$ from 1 to $L$, $i$ from 1 to $m$,}
       \STATE{Identify the pattern $t^*$ to which $\q$ corresponds for the $i\thup$ $\ell$-tuple of hyperplanes.}
	\FOR{$g$ from $1$ to $G$}
		\STATE{Update $\tilde \r_k(g)=\tilde \r_k(g)+\r_k(\ell,i,t^*,g)$.}
	\ENDFOR
	\ENDFOR
	\STATE{Set $\q = \sign(\A^{(k)} \tilde \r_k)$.}
	\ENDFOR
	\STATE{Classify $ b=\argmax_{g\in\{1,\cdots, G\}} \tilde \r_K(g)$.} 
\end{algorithmic}    
\end{algorithm}

\section{Experimental results} \label{sec:exper}
We test ISCB on synthetic and image datasets. The synthetic datasets demonstrate why the iterative method is effective for certain simple settings and how the data transforms between iterations. ISCB is also tested on the MNIST dataset of hand-written digits \cite{MNIST}, the YaleB dataset for facial recognition \cite{YaleB1,YaleB2,YaleB3,YaleB4} and the Norb dataset for classification of images of various toys \cite{Norb}.

\subsection{Two-dimensional synthetic data}
\begin{figure}
\centering
\includegraphics[width = 0.35\textwidth]{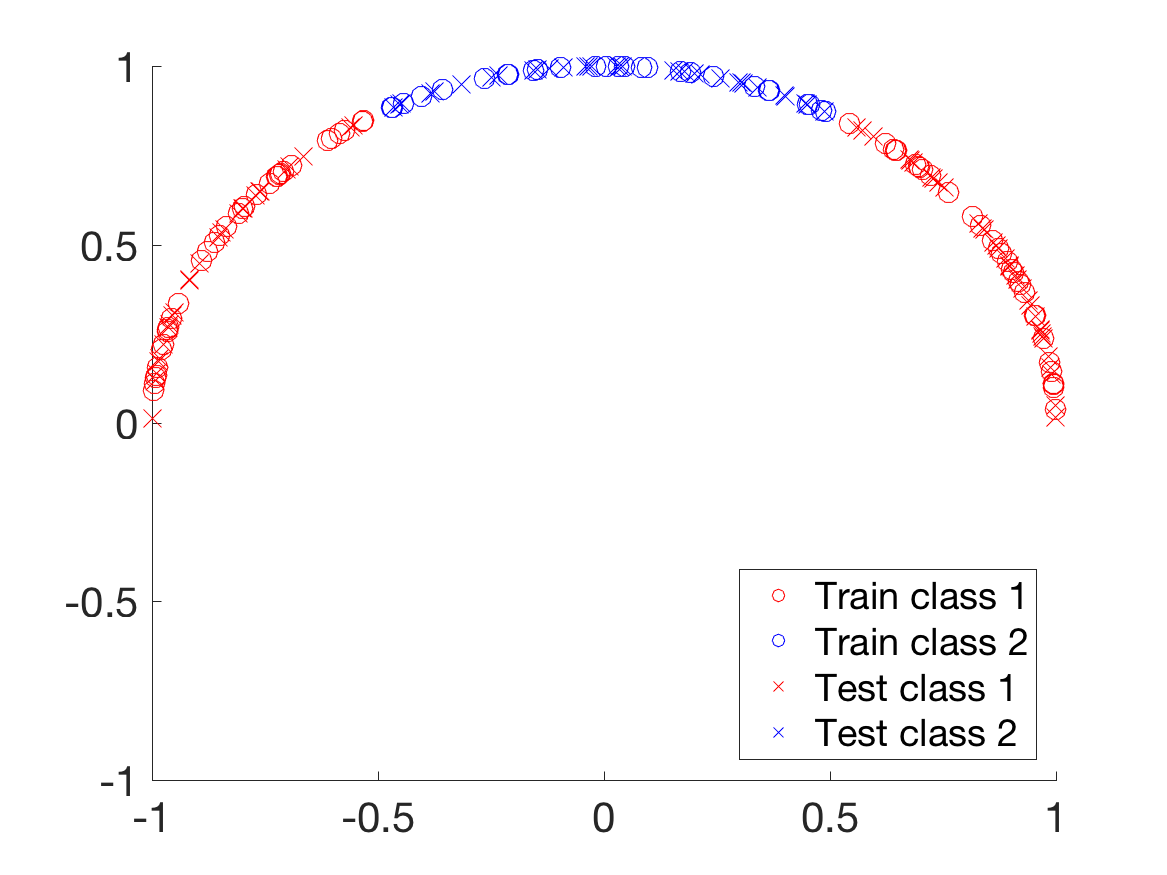}
\includegraphics[width = 0.35\textwidth]{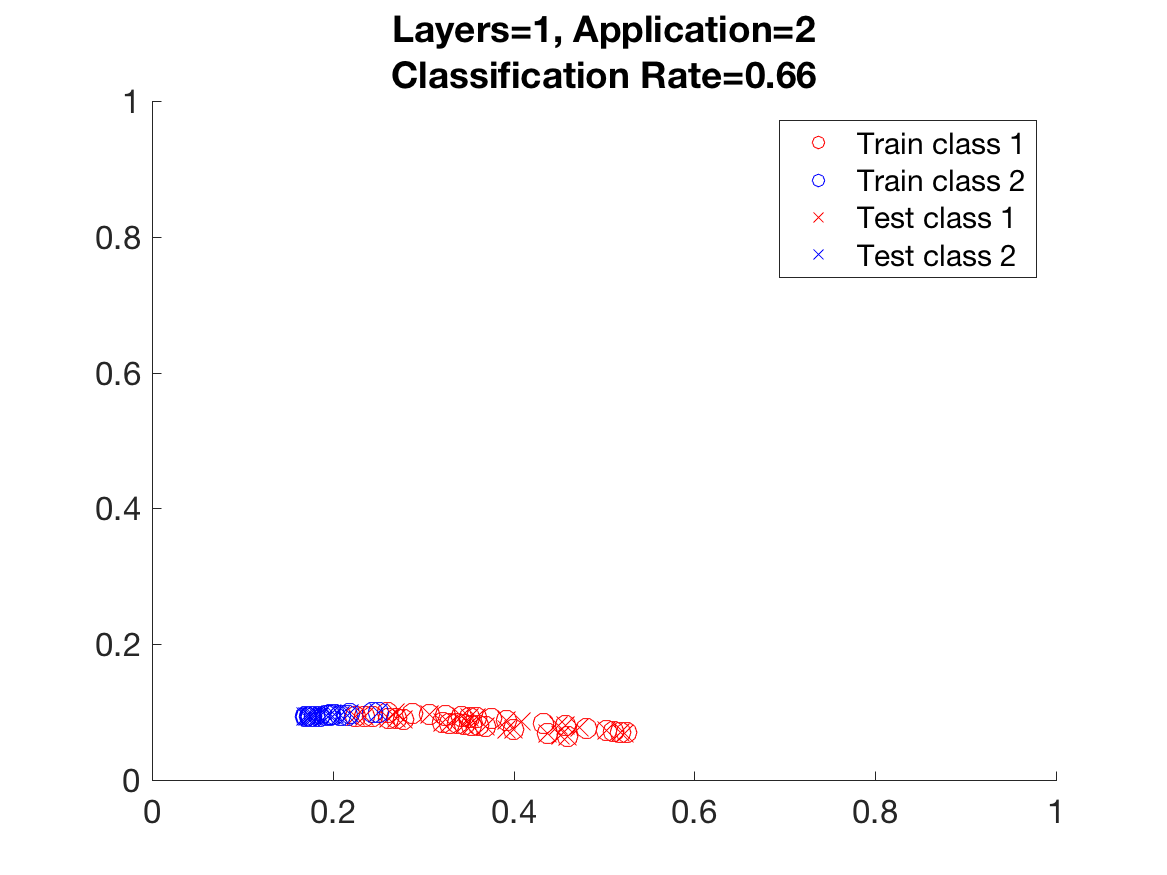}
\includegraphics[width = 0.35\textwidth]{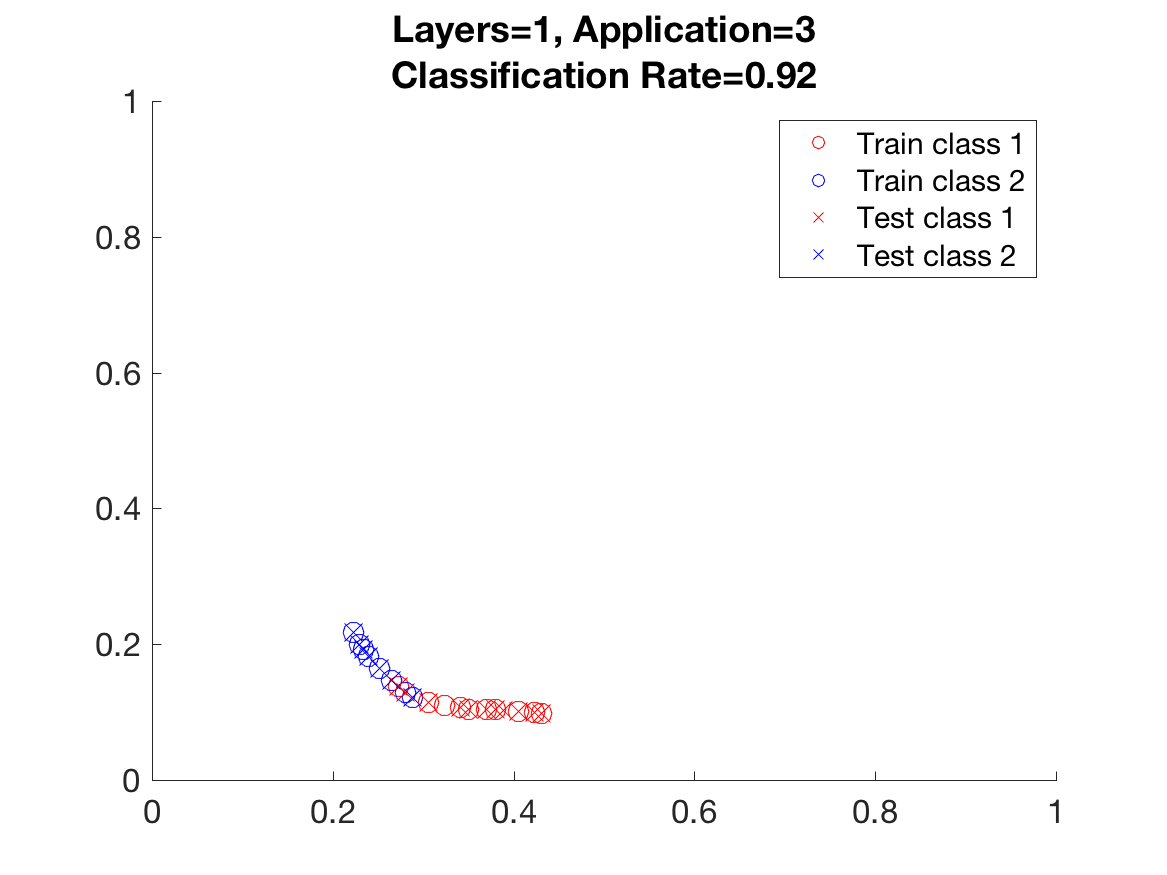}
\includegraphics[width = 0.35\textwidth]{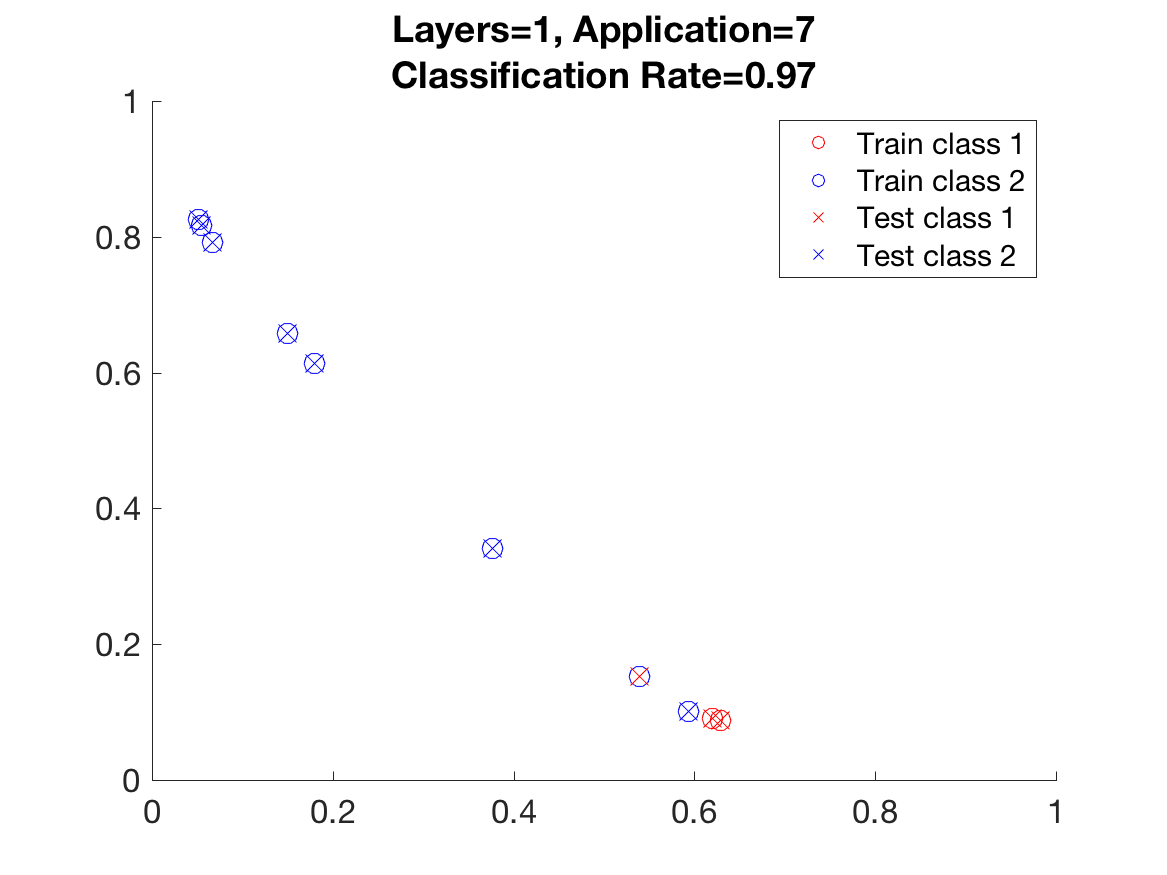}
\caption{The $\tilde \r_k$ vectors are plotted for ISCB with varying numbers of iterations $k$. The original training and testing data are shown in the upper left plot. Circles indicate training data and crosses indicate testing data. One level is used for each application of \SCB\ and the subsequent plots give the $\tilde \r_k$ vectors for $k = 1, 3,$ and $7$ respectively. The classification rate after a single application of SCB is 66\%. After three iterations the classification rate is 92\% and after seven application the classification rate is 97\%. 
}
\label{fig:dataprogr}
\end{figure}

We further motivate ISCB through examples with two-dimensional synthetic data. For two-dimensional data with two classes, the dimension of the input data for all applications of \SCB\ is two-dimensional and so we can easily visualize the training and testing data at each iteration. Consider the data given in the upper left plot of \Cref{fig:dataprogr}. There are two times as many points from the red class considered, both in the training and testing set. Half of the red points in testing and training lie on either side of the blue data points. Thus, applying \SCB\ using a single level leads to all of the blue test points being misclassified as red. The abhorrent misclassification of the blue points is caused by the fact that we are using only a single level ($L=1$) and for any hyperplane at least as many red points as blue lie on either side of it.  
The $\tilde \r_1$ values, plotted in the upper right plot of \Cref{fig:dataprogr}, have a much nicer distribution in terms of ease of classification; in fact, they are nearly linearly separable. The separation in the $\tilde \r_1$ values between the blue and red points occurs since $\tilde \r_1(\text{red})$ is generally larger for red points than for the misclassified blue points. That is, the points that truly belong to the red class are more ``confidently" classified as red than are the blue points. If we consider the $\tilde \r_1$ values as data, applying \SCB\ now classifies the data with much higher accuracy (92\% as compared to 66\% for the original training data), while still only using a single level. By the seventh iterative application of \SCB, the accuracy increases to 97\%. If we perform the same experiment, but include a higher density of blue points so that the total number of red and blue points are the same, we achieve higher accuracy at the first application of \SCB, but again see improved accuracy for later iterations.

•

\subsection{Image datasets}\label{sec:MNIST}
\begin{figure}
\centering
\includegraphics[width = 0.32\textwidth]{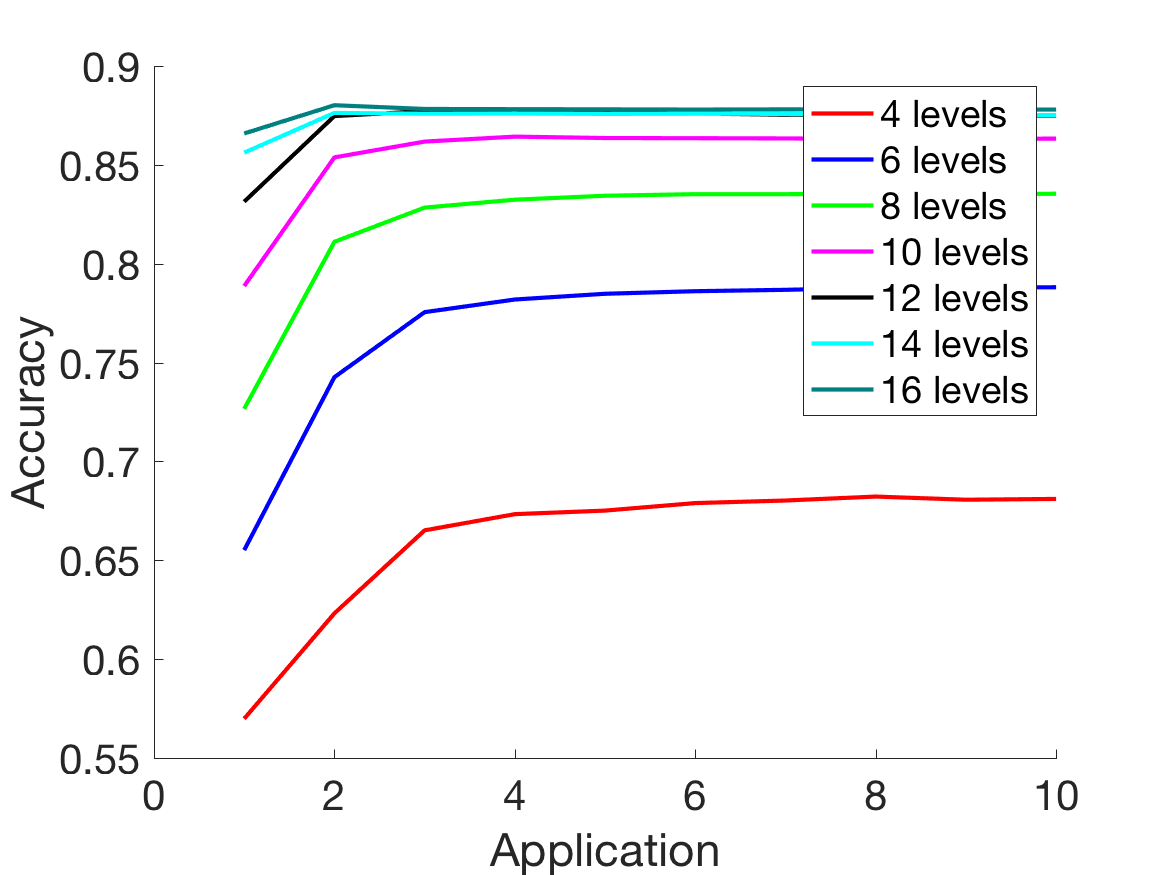}
\includegraphics[width=0.32\textwidth]{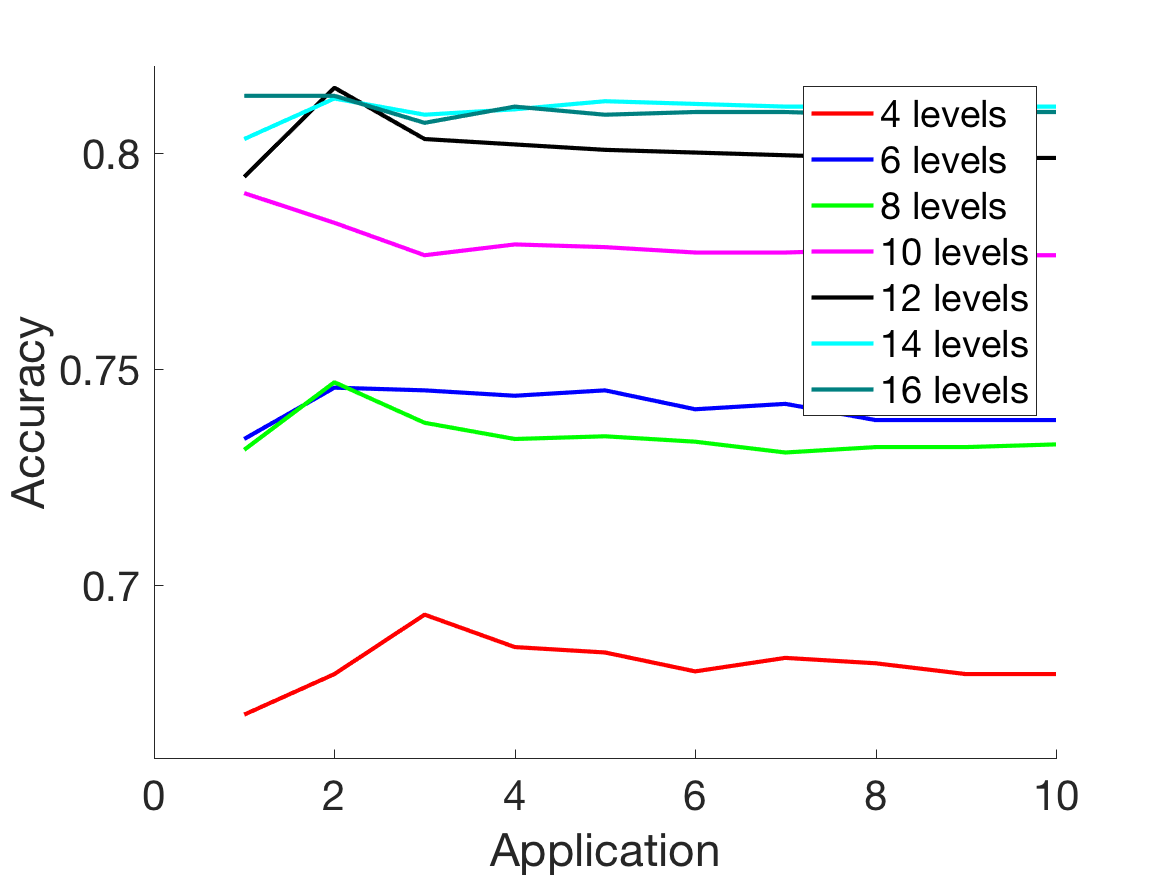}
\includegraphics[width=0.32\textwidth]{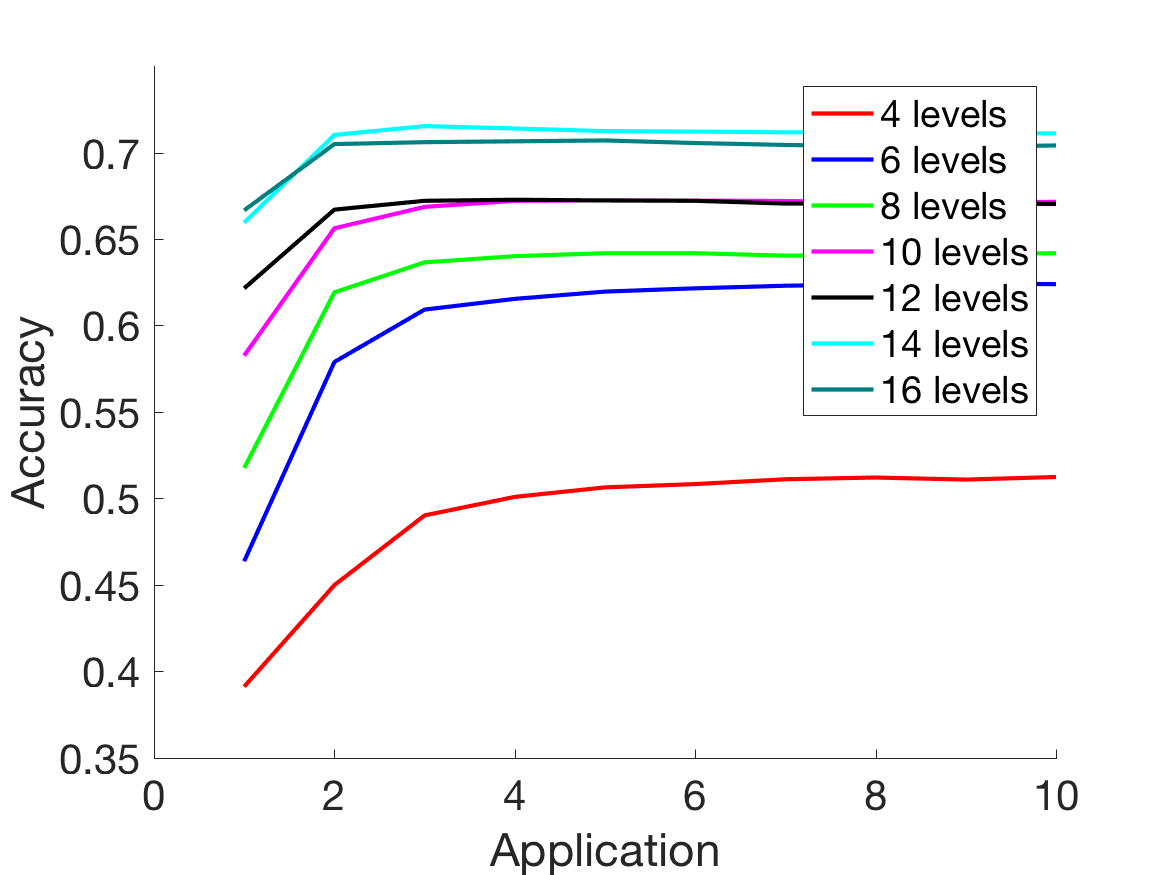}
\caption{Accuracies for classifying MNIST data among 10 classes (left plot), YaleB data among eight classes (middle plot) and Norb data among five classes (right plot) are given in terms of the number of applications of \SCB\ used. For the MNIST dataset, the model is trained with $p=1000$ images of each class and tested on 100 images from each class. The model for the YaleB dataset is trained using $p=40$ training images from each class and applied to 20 test images from each class. For the Norb dataset, the model is trained on $p=1000$ training images and is applied to 200 test images for each class.
In each model, $m=500$ measurements are used. Results are averaged over 10 trials.  }
\label{fig:perf}
\end{figure}

We test ISCB on the MNIST dataset of hand-written digits \cite{MNIST}, the YaleB dataset for facial recognition \cite{YaleB1,YaleB2,YaleB3,YaleB4}, and the Norb dataset for classification of images of various toys \cite{Norb}. Results are shown in \Cref{fig:perf}.  
We generally find both that increasing the number of levels used in each \SCB\ application of ISCB and increasing the number of applications leads to improved performance. The classification accuracies typically level off after only a few applications of \SCB, with the largest improvement typically occuring between the first and second application. These trends are less clear in the YaleB dataset, but this may be in part due to the limited amount of training data available for this dataset.  

•

\section{Alternative iterative method}\label{sec:rlg}

We motivated ISCB by noting that the $\tilde \r$ vectors from \Cref{algo:Class} for test data from the same classes share similar structures. We additionally find that the contributions to the $\tilde \r$ values coming from different levels 
 admit different patterns as well. We could thus choose to use 
\[\hat \r_k(\ell,g) = \sum_{i=1}^m \sum_{t^*} \r_k(\ell,i,t^*,g)\]
 as data for the $k\thup$ application of \SCB\ instead of $\tilde \r_k(g)$ as is done in \Cref{algo:ItTrain}. Here, $t^*$ ranges over all observed sign patterns for the $i\thup$ $\ell$-tuple of hyperplanes. We refer to this method as \textit{ISCB with $\hat \r$}. Note that we have the following relation between $\tilde \r_k$ and $\hat \r_k$,
\[\tilde \r_k(g) = \sum_{\ell = 1}^L \hat \r_k(\ell,g).\]
After the first application of \SCB, the dimension of the data for ISCB with $\hat \r$ is now $\RR^{L  G}$.

\begin{figure}
\centering
\includegraphics[width=0.45\textwidth]{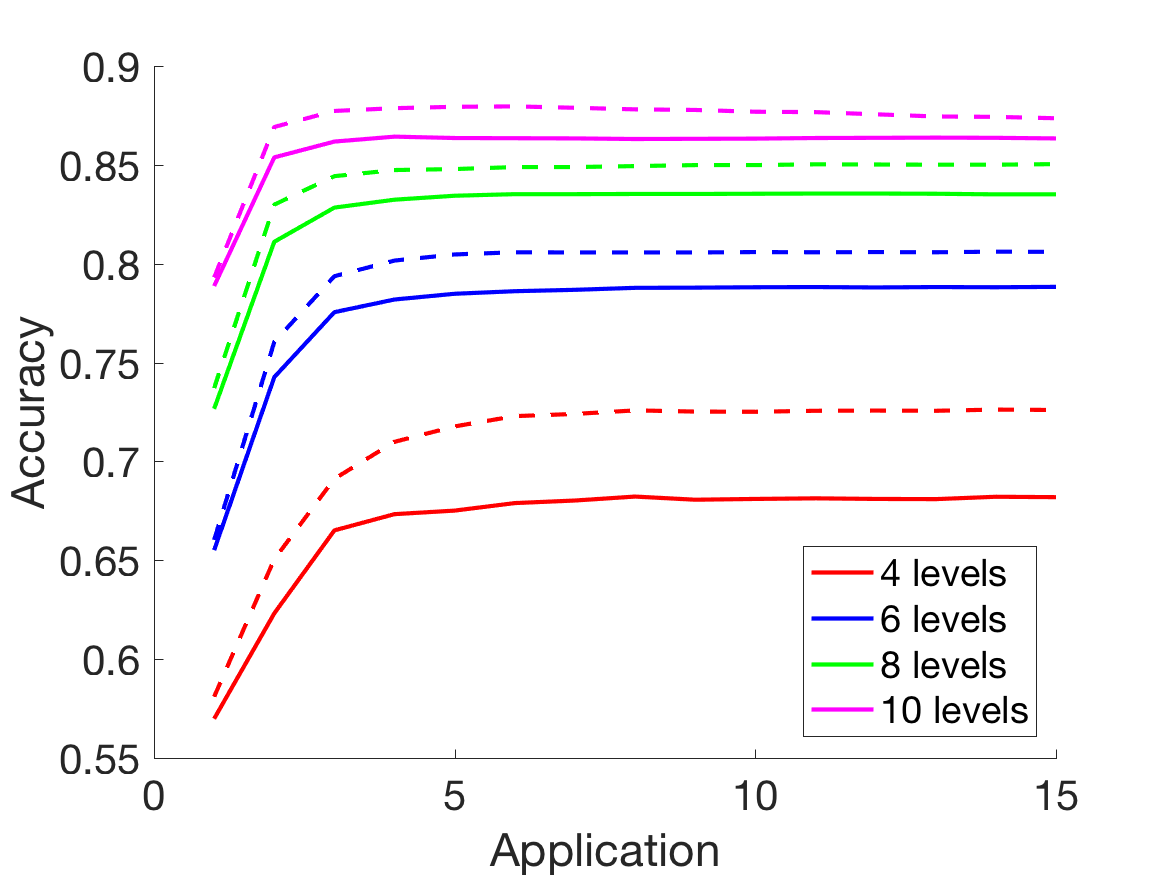}
\includegraphics[width=0.45\textwidth]{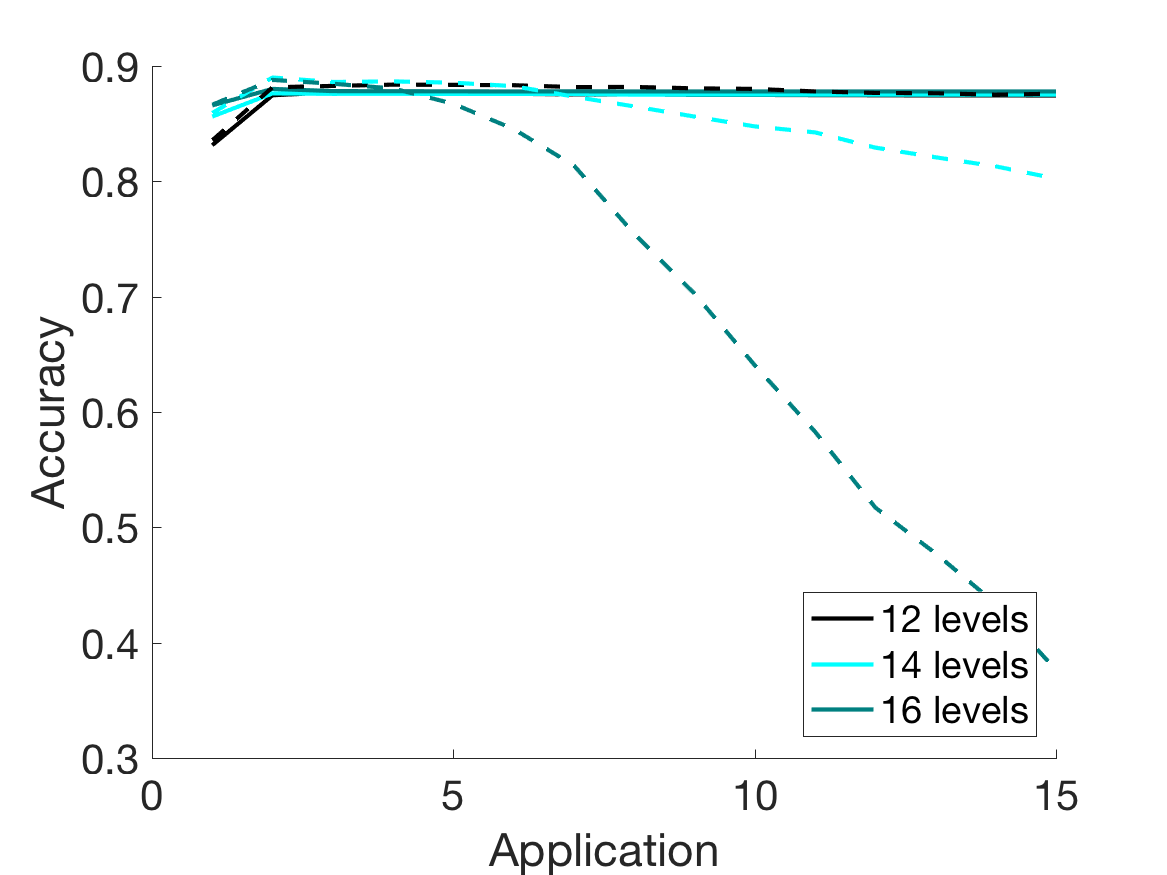}
\caption{The performance of ISCB using $\tilde \r$, presented in \Cref{sec:iter_alg}, (solid) is compared with that of the alternative version  of ISCB using $\hat \r$, presented in \Cref{sec:rlg}, (dashed) on the MNIST dataset. $p=1000$ training and 100 testing images are used for each digit. Each method uses $m=500$ binary measurements of the data at each application of \Cref{algo:Train}. The number of levels $L$ used with each method is indicated in the legend. }
\label{fig:rlg_v_rg}
\end{figure}

In certain settings, ISCB with $\hat \r$ performs better than ISCB of \Cref{sec:iter_alg}. 
Typically, using $\hat \r_k(\ell,g)$  as opposed to $\tilde \r_k(g)$ as input to the subsequent applications of \Cref{algo:Train} performs better when the number of levels $L$ used is small.   
Unfortunately, for higher numbers of levels $L$ we see drastic declines in performance for later applications when using $\hat \r_k(\ell,g)$, as this method is more prone to overfit. These trends are illustrated in \Cref{fig:rlg_v_rg} for the MNIST dataset. In the left plot of \Cref{fig:rlg_v_rg}, ISCB with $\hat \r$ leads to improved performance over ISCB with $\tilde \r$. As the number of levels $L$ used increases from four to 10, however, this difference diminishes. For greater than 14 levels, using ISCB with $\hat \r_k(\ell,g)$ leads to decreasing performance in the number of applications of \SCB\ (seen in the right plot of \Cref{fig:rlg_v_rg}). The same decrease in performance does not occur when using the $\tilde \r_k(g)$ values as data for the next iteration.

•

\subsection{Overfitting}
ISCB is relatively prone to overfitting, since the model output for training and testing data at iteration $k$ are used as input for the $(k+1)\stup$ iteration. Thus, any overfitting that occurs at earlier iterations gets propagated to later iterations. In particular, if we overfit to the training data at iteration $k$, then the training and testing data at the next iteration are no longer sampled from similar distributions. ISCB with $\hat \r$ has a much greater propensity to overfit as compared to ISCB with $\tilde \r$, especially when the number of levels is large. This effect makes intuitive sense since for longer $\ell$-tuples of hyperplanes the testing data is less likely to match the sign patterns of training data and so the $\hat \r_k(\ell,g)$ values for training and testing data may diverge for longer $\ell$-tuples and later iterations $k$.
These observations suggest that choosing an appropriate number of levels is especially critical for ISCB as compared to \SCB. Fortunately, if a model is trained using too many levels, one could simply 
use the model output from the first application to arrive at more accurate predictions. In particular, there is no need to re-train the model.

\section{Theoretical analysis}\label{sec:theory}
We next offer some theoretical analyses pertaining to why we expect performance to improve through multiple applications of SCB for several simple scenarios. At a high level, the iterative framework has the opportunity to train on its own output and correct misclassifications that occur in previous iterations. Qualitatively, as the number of iterations increases, we find that the data points that are more easily identifiable as belonging to a single class are pushed toward extreme points of the range of outputs, while data points that are more difficult to classify fall in the interior of the range and have the chance to be classified correctly at the next iteration.

\subsection{Binary classification of point masses of equal mass}

As a first simple but illustrative example, consider a classification task between two classes. 
Assume that the training and testing data for each class is concentrated at a single point, i.e. a point mass, and that each class has the same number of training points or equivalently that each point mass has the same density. 
Let $j$ be the number of hyperplanes that separate the two point masses at the first application of the classification method ($j$ will clearly depend on the angle between the point masses and can be easily bounded probabilistically).
Consider the simplified setting in which we use a single level;  
note that in this case, since $L=1$, we have $\tilde \r_k = \hat \r_k$ and so ISCB and ISCB with $\hat \r$ are equivalent.
With this setup, for testing data in class 1,
\[\tilde \r_1 = \left(\sum_{i=1}^m \r(\ell,i,t^*,1),\sum_{i=1}^m \r(\ell,i,t^*,2) \right) = (j,0). \]
For testing data in class 2,
\[\tilde \r_1 = \left(\sum_{i=1}^m \r(\ell,i,t^*,1),\sum_{i=1}^m \r(\ell,i,t^*,2) \right) = (0,j). \]
Thus, if at least one hyperplane separates the two point masses initially, then at the next iteration, the angle between data points of class 1 and 2 is $\pi/2$ (the best possible). Since the data are two-dimensional and we restrict the hyperplanes to intersect the positive quadrant after the first \SCB\ application, then if the model classifies the point masses correctly at the first iteration, it will correctly classify at all subsequent iterations as well.

\subsection{Binary classification of point masses}

\begin{figure}
\centering
\includegraphics[width = .5\textwidth]{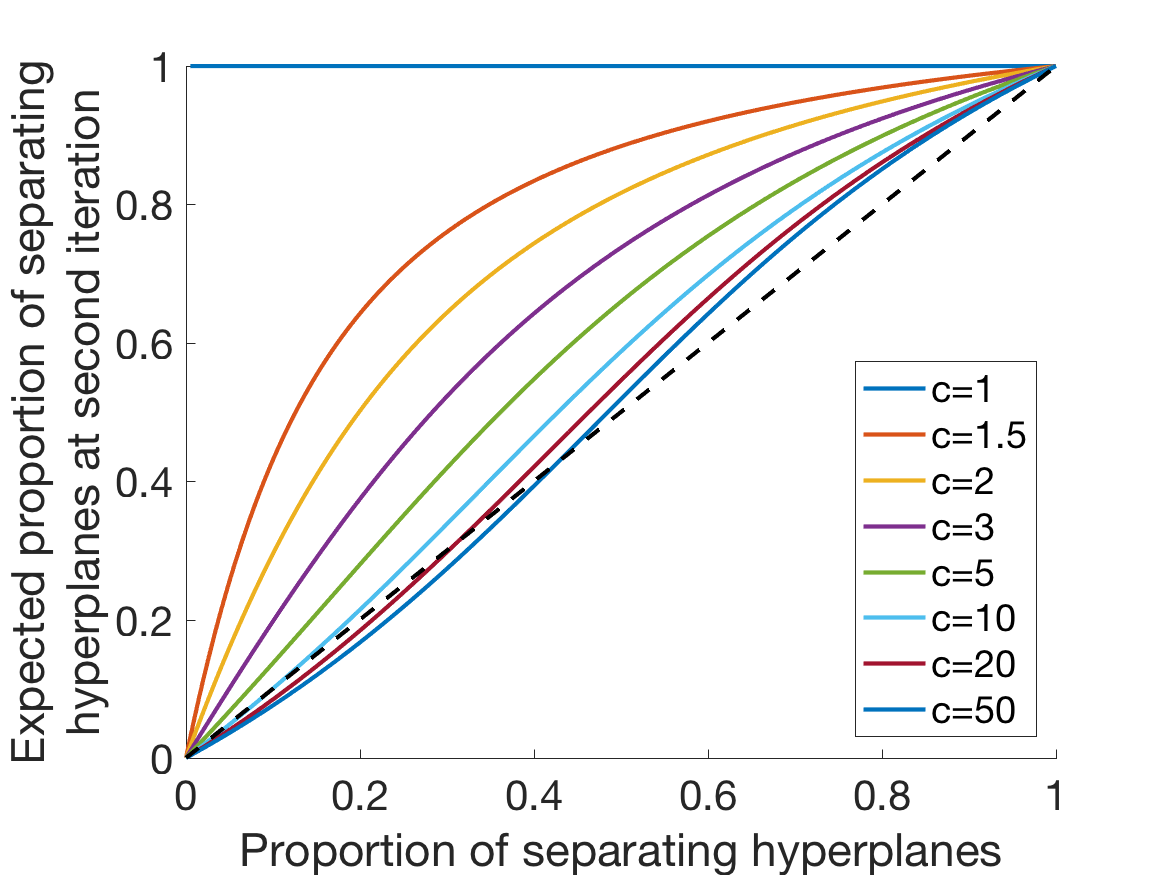}
\caption{This plot shows the expected proportion of hyperplanes that separate data at the second iteration of ISCB given the fraction of separating hyperplanes at the first application of \SCB. The relative sizes of the two classes, given by $A_1$ and $A_2$, are varied as well, as is indicated by the parameter $c = \frac{A_1}{A_2}$ given in the legend. }
\label{fig:sepAngle}
\end{figure}

We next consider the slightly more involved setting in which the data from each class is again concentrated at a single point, however, the number of points in the two classes differ. We again consider only a single level $L$ and let $j$ be the number of hyperplanes that separate the two point masses in the first application of \SCB. In expectation, $\frac{j}{m}$ gives an indication of the angle separating the two point masses, where $m$ is the number of rows in the measurement matrix. Let $A_1$ be the number of points in class 1 and $A_2$ be the number of points in class 2.  
For testing data in class 1,
\begin{align*}
\tilde \r_1(1) &= \sum_{i=1}^m \r(\ell,i,t^*,1) = j+(m-j)\frac{A_1|A_1-A_2|}{(A_1+A_2)^2} \text{ and }\\
 \tilde \r_2(2) &= \sum_{i=1}^m \r(\ell,i,t^*,2) = (m-j)\frac{A_2|A_1-A_2|}{(A_1+A_2)^2}.
\end{align*}
For testing data in class 2,
\begin{align*}
\tilde \r_1(1) &= \sum_{i=1}^m \r(\ell,i,t^*,1) = (m-j)\frac{A_1|A_1-A_2|}{(A_1+A_2)^2}\text{ and }\\
\tilde \r_2(2) &= \sum_{i=1}^m \r(\ell,i,t^*,2) =j+ (m-j)\frac{A_2|A_1-A_2|}{(A_1+A_2)^2}.
\end{align*}
Note that the data for the second application of the method are again two-dimensional. Let $\tilde \g_1$ be the $\tilde \r_1$ vector for a data point in class 1 and $\tilde \g_2 $ be the $\tilde \r_1$ vector for a data point in class 2. The following formula gives the angle $\theta$ between the two point masses at the second application,
\begin{equation}\theta = \cos^{-1}\left( \frac{\langle \tilde \g_1, \tilde \g_2\rangle}{||\tilde \g_1||_2 \cdot ||\tilde \g_2||_2}\right).
\label{eqn:angle}
\end{equation}
\Cref{fig:sepAngle} shows the angle that separates the point masses of the training data at the second application in terms of $\frac{j}{m}$ for various ratios $c = \frac{A_1}{A_2}$. We find that if $A_1$ and $A_2$ are similar in size, then the expected angle separating the two point masses increases for the second application, making the point masses ``easier" to separate in later applications.

•

\subsection{Symmetric data} 
\label{sec:SymmData}

We consider another simple, two-dimensional, two-class setting.  
 Assume that the training data for the two classes lie in the positive quadrant of the plane and are distributed symmetrically about the line $y=x$, as are the $m$ hyperplanes. If we assume that the hyperplanes are uniformly distributed, this is a reasonable simplification. We could also alternatively enforce this condition when generating the random hyperplanes, although a generalization of this strategy to higher dimensions is not immediate. We show that if we have at least one hyperplane that separates the two classes, then the data will be classified correctly via \SCB\ with a single level. Although this statement provides a classification guarantee for \SCB, the data structure is representative of $\tilde \r_k$ values for binary classification in which the previous iteration of ISCB has classified all of the data points correctly. In this sense, this result can be interpreted as providing a guarantee that if ISCB has performed well at the previous iteration, we should expect it will perform well at the next iteration also. 
Let $n$ be the number of data points in each class. For simplicity, we will refer to the class that lies above the line $y=x$ as class 1 and the class that lies below the line $y=x$ as class 2. 

\begin{figure}
\centering
\includegraphics[width=0.6\textwidth]{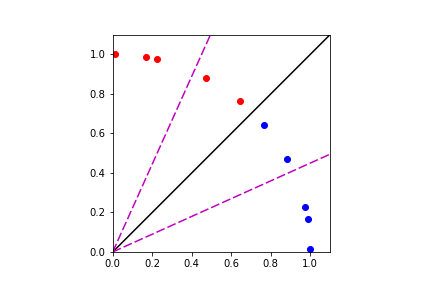}
\caption{Illustration of data setup for \Cref{sec:SymmData}. Red and blue points indicate data from class 1 and 2 respectively. The dashed purple lines provide an example pair of hyperplanes. We assume the data are distributed symmetrically about the black line $y=x$ as are the hyperplanes.}
\label{fig:SymmData}
\end{figure}

Consider a hyperplane that intersects the region above the line $y=x$, for example, the purple dotted hyperplane in \Cref{fig:SymmData}. Let $s$ be the number of points belonging to class 1 that lie below this hyperplane. Each such hyperplane that lies above the line $y=x$ has a corresponding hyperplane below the line $y=x$ by the assumed symmetry condition. Then $s$ also gives the number of points in class 2 that lie above this corresponding hyperplane. 
Contributions to the $\tilde \r$ vector are summarized in \Cref{tab:symmTheory} for hyperplane pairs, i.e. hyperplanes that are symmetric about the line $y=x.$

\begin{table}[ht]
\begin{center}
\def\arraystretch{1.25}
\begin{tabular}{|c|c c | c c|}
\hline 
Hyperplane & Class 1 & & Class 2&  \\ \hline
 &+&-&+&-\\\hline
Lies above $y=x$ & 1& $\frac{s}{n+s}\frac{n-s}{n+s}$ & 0 & $\frac{n}{n+s}\frac{n-s}{n+s}$\\ \hline
Lies below $y=x$  & $\frac{n}{n+s}\frac{n-s}{n+s}$ & 0 & $\frac{s}{n+s}\frac{n-s}{n+s}$ & 1\\ \hline
\end{tabular}•
\end{center}
\caption{Contributions to the $\tilde \r$ vector corresponding to a data point $p$ and for a hyperplane pair, which are symmetric about the line $y=x$. The symbols $+$ and $-$ indicate that the point $p$ lies above and below the hyperplane, respectively.}
\label{tab:symmTheory}
\end{table}

Consider now a test point $\x$ from class 1. 
Suppose that $j$ hyperplanes cut between the test point $\x$ and the line $y=x$. Let $s_i$ be the $s$ value for the $i\thup$ hyperplane. Then at the next application, 
\[\tilde \r(1) = j + \sum_{i=1}^{m/2-j} \frac{s_i}{n+s_i} \frac{n-s_i}{n+s_i} + \sum_{i=1}^{m/2} \frac{n}{n+s_i} \frac{n-s_i}{n+s_i} \]
and
\[\tilde \r(2) = \sum_{i=1}^{m/2-j} \frac{n}{n+s_i} \frac{n-s_i}{n+s_i} + \sum_{i=1}^{m/2} \frac{s_i}{n+s_i} \frac{n-s_i}{n+s_i} .\]
The first summation in each equation gives the contribution to the $\tilde \r$ vector from hyperplanes that lie above the line $y=x$, but below the data point $\x$. The second summation gives the contribution from hyperplanes that lie below $y=x$. Each $s_i$ in the first summation corresponds to an $s_i$ in the second summation. 

Subtracting these values,
\[\tilde \r(1) -\tilde \r(2) = j - \sum_{i=1}^{m/2-j} \frac{n-s_i}{n+s_i} \frac{n-s_i}{n+s_i} + \sum_{i=1}^{m/2} \frac{n-s_i}{n+s_i} \frac{n-s_i}{n+s_i}  = j+  \sum_{i=1}^{j} \left(\frac{n-s_i}{n+s_i} \right)^2 \ge0 .\]
 Thus, we find that the test point $\x$ will always be classified correctly, in fact, by at least a margin of $j$. 
(If there are no training points from class 1 between the test point of class 1 and the training points of class 2, the $\tilde \r$ values for the two classes will be equal if $j=0$.) 
Since the data is then perfectly classified in this application of \SCB\ and the resulting $\tilde \r$ vectors will again be symmetric about the line $y=x$, if we make the same symmetry assumptions about the hyperplanes, we can apply this same result to future applications of \SCB\ with a single level. Thus, for this simple symmetric data, ISCB will classify the data perfectly using a single level ($L=1$) and for any number of iterations $K$.

\subsection{Probabilistic bounds for an angular model}
\label{sec:Thm1Theory}

\begin{figure}
\centering
\includegraphics[width=0.4\textwidth]{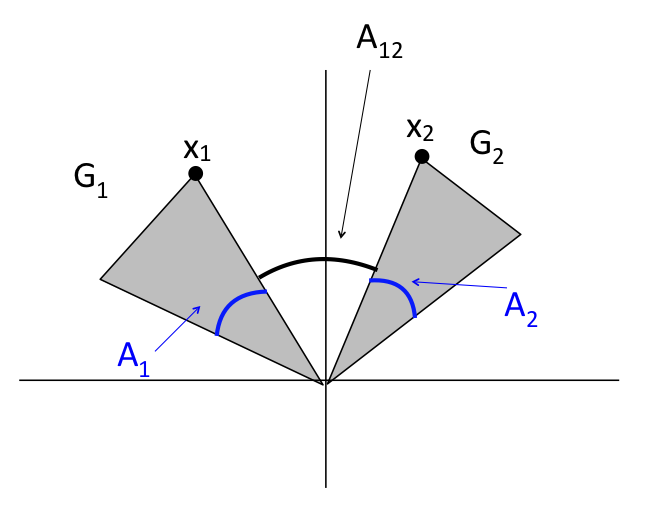}
\caption{Illustration of data setup for \Cref{sec:Thm1Theory}.}
\label{fig:setup}
\end{figure}

We next consider an analogue to Theorem 1 of \cite{NSW17Simple}, although the setting is modified slightly. Consider two-dimensional data with two classes. Suppose that the data from each class is distributed within the disjoint wedges, $G_1$ and $G_2$, with angles $A_1$ and $A_2$ respectively. This setup is  illustrated in \Cref{fig:setup}.
Consider the data points $\x_1$ and $\x_2$, which lie on the inside edge of each wedge. Let $A_{12}$ be the angle between these two points. We aim to find a lower bound on the angle between the $\tilde \r_1$ vectors for $\x_1$ and $\x_2$ after a single application of \SCB\ with a single level $L$. Again, since we only use a single level, $\hat \r_1= \tilde \r_1$ for all points $\x$.

\begin{table}
\begin{center}
	\def\arraystretch{1.5}
  \begin{tabular}{ | c | c | c | c |}
    \hline
   Hyperplane case & Number in event  & Class & Value of $\r(1,i,t,g)$\\ \hline
    Separates $\x_1$ and $\x_2$ & $j$  & 1 & $1$\\ 
	& & 2 & $0$ \\ \hline
    Does not separate $\x_1$ and $\x_2$  & $m - j-k_1-k_2$ & 1 & $\frac{A_1|A_1-A_2|}{(A_1+A_2)^2}$\\ 
	or intersect $G_1$ or $G_2$ & & 2 & $\frac{A_2|A_1-A_2|}{(A_1+A_2)^2}$ \\ \hline 
    Intersects $G_2$ & $k_2$ & 1 & $\frac{A_1|A_1-A_2u'|}{(A_1+A_2u')^2}$\\ 
	& & 2 & $\frac{A_2u'|A_1-A_2u'|}{(A_1+A_2u')^2}$ \\ \hline  
    Intersects $G_1$ & $k_1$  & 1 & $\frac{A_1u|A_1u-A_2|}{(A_1u+A_2)^2}$\\ 
	& & 2 & $\frac{A_2|A_1u-A_2|}{(A_1u+A_2)^2}$\\ \hline
  \end{tabular}
\end{center}
\caption{Contributions to the membership index parameter $\r$ for the point $\x_1$ and for hyperplanes of various types. The variables $u$ and $u'$ are i.i.d. random variables uniformly distributed between zero and one, indicating the angle at which random hyperplanes intersect the wedges $G_1$ and $G_2$. $A_1, A_2,G_1,G_2,\x_1$ and $\x_2$ are as shown in \Cref{fig:setup}.} 
\label{tab:r_contr}
\end{table} 

Assume that the data is distributed uniformly in $G_1$ and $G_2$. 
Let $k_1$ and $k_2$ be the number of hyperplanes that intersect $G_1$ and $G_2$ respectively and let $j$ be the number of hyperplanes that separate $G_1$ and $G_2$. Note that 
\[\E k_1 = \frac{A_1}{\pi},\quad \E j = \frac{A_{12}}{\pi},\quad \text{and}\quad \E k_2 = \frac{A_2}{\pi}.\] Assume that the hyperplanes are also distributed with uniformly random angles within these wedges. We can then replace $P_{g|t}$ with angular measures, specifically, $A_i u_h$, where $u_h\in[0,1]$ and depends on the angle at which the hyperplane $h$ intersects $G_i$. Since the hyperplanes are uniformly distributed at random within each region, the $u_h$ are uniform random variables between zero and one. 

\begin{figure}
\centering
\includegraphics[width = 0.4\textwidth]{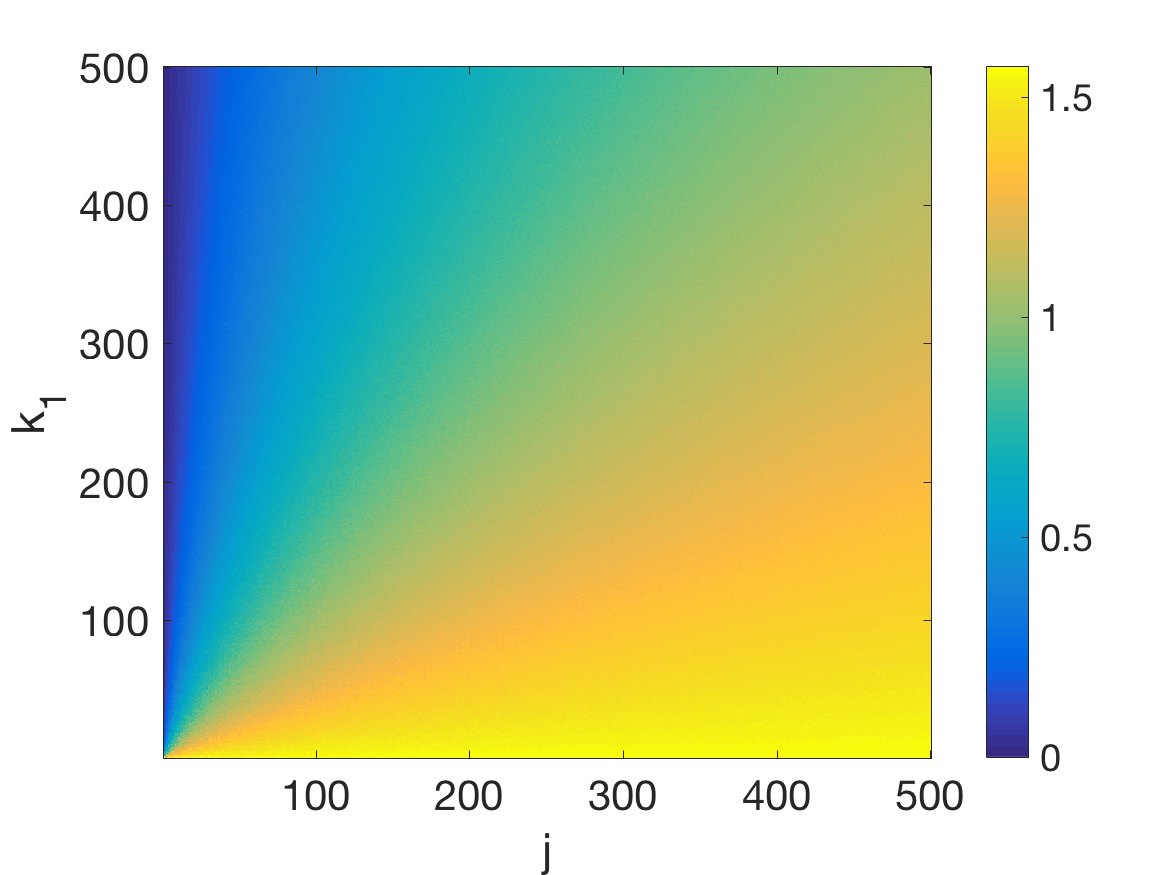}
\includegraphics[width = 0.4\textwidth]{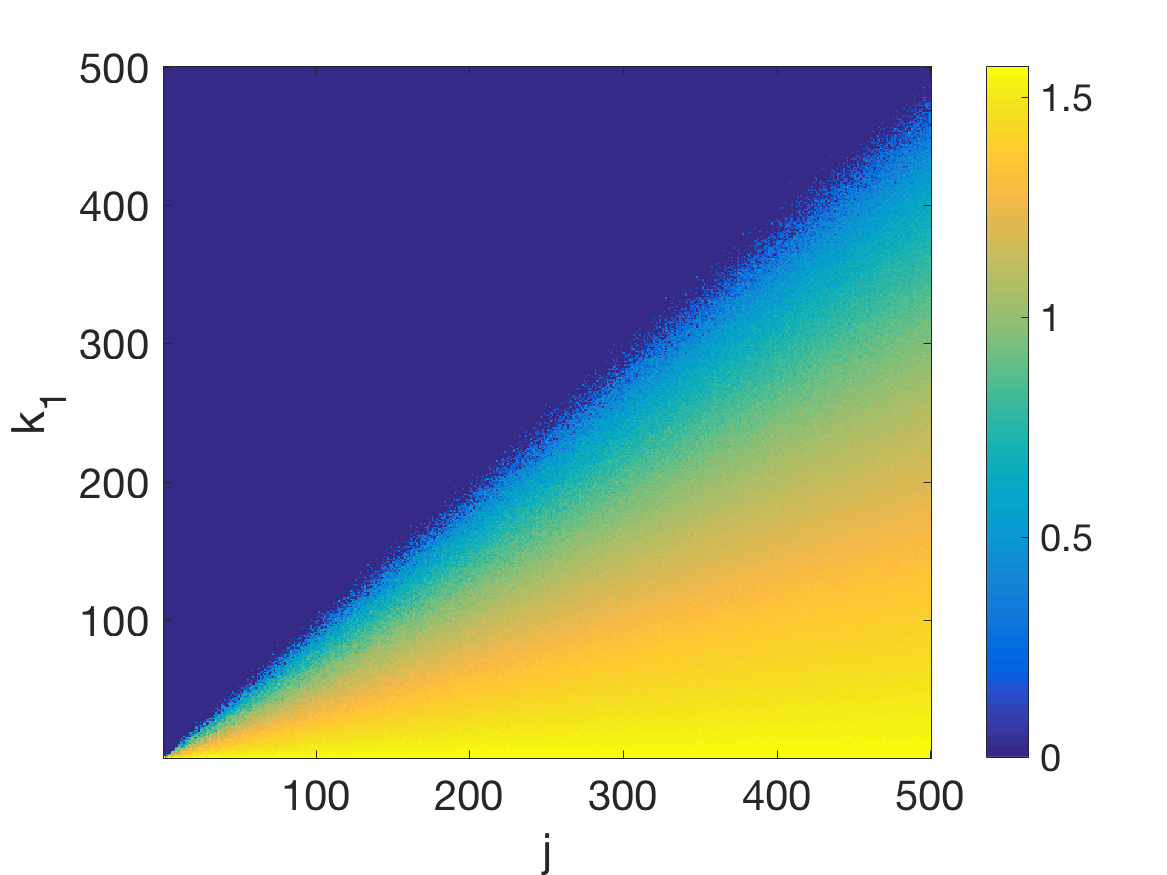}
\caption{For various values of $k_1=k_2$ (the number of hyperplanes intersecting the wedges $G_1$ and $G_2$ respectively) and $j$ (the number of hyperplanes separating the wedges $G_1$ and $G_2$), the left plot indicates the true angle (in radians) between $\tilde \g_1$ and $\tilde \g_2$ as given in \Cref{eqn:gTilde1,eqn:gTilde2}. 
The right plot indicates the angle using the upper bound for $\cos(\theta)$ given in \Cref{eqn:angleBound}.  }
\label{fig:boundComp}
\end{figure}

 The contribution to the membership index parameter $\r$ for \SCB\ with a single level $L$ and for each possible type of hyperplane in this setup for the point $\x_1$ are summarized in \Cref{tab:r_contr}. 
To simplify calculations, assume that $A_1 = A_2$. With this assumption, the membership index parameters no longer depend on $A_1$ or $A_2$. Summing over all hyperplanes, for $\x_1$ we have
\begin{align*}
\tilde \r_1(1) = \sum_{i=1}^m \r(1,i,t_i^*,1) 
&= j+\sum_{h=1}^{k_1}\frac{u_h(1-u_h)}{(u_h+1)^2}
+ \sum_{h=1}^{k_2}\frac{1-u_h'}{(1+u_h')^2}
\end{align*}
and 
\begin{align*}
\tilde \r_1(2) =\sum_{i=1}^m \r(1,i,t_i^*,2) 
&=   \sum_{h=1}^{k_{1}}\frac{1-u_h}{(u_h+1)^2} + \sum_{h=1}^{k_2}\frac{u_h'(1+u_h')}{(1+u_h')^2}.
\end{align*} 
The calculation for $\x_2$ is similar. Let $\tilde \g_1$ and $\tilde \g_2$ be the $\tilde \r_1$ vectors corresponding to $\x_1$ and $\x_2$ respectively. Then 
at the next application, we have 
\begin{equation} 
\tilde \g_1 = \left(j+\sum_{h=1}^{k_1}\frac{u_h(1-u_h)}{(u_h+1)^2}
+ \sum_{h=1}^{k_2}\frac{1-u_h'}{(1+u_h')^2}, \sum_{h=1}^{k_{1}}\frac{1-u_h}{(u_h+1)^2} + \sum_{h=1}^{k_2}\frac{u_h'(1-u_h')}{(1+u_h')^2}\right)
\label{eqn:gTilde1}
\end{equation}
and 
\begin{equation}
\tilde \g_2 = \left(\sum_{h=1}^{k_{1}}\frac{u_h(1-u_h)}{(u_h+1)^2} + \sum_{h=1}^{k_2}\frac{1-u_h'}{(1+u_h')^2} ,j+\sum_{h=1}^{k_1}\frac{1-u_h}{(u_h+1)^2}
+ \sum_{h=1}^{k_2}\frac{u_h'(1-u_h')}{(1+u_h')^2}\right).
\label{eqn:gTilde2}
\end{equation}
The angle between these two vectors is again given by \Cref{eqn:angle}. The resulting angles from simulations for various $k_1=k_2,$ and $j$ are given in the left plot of \Cref{fig:boundComp}. We make the simplification $k_1=k_2$ to ease visualization. Unsurprisingly, as $j$ increases so does the separation between $\tilde \g_1$ and $\tilde \g_2$ at the second iteration. As $k_1$ and $k_2$ increase, for fixed $j$, the separation between the two points at the next application decreases.

\begin{figure}
\centering
\includegraphics[width = 0.3\textwidth]{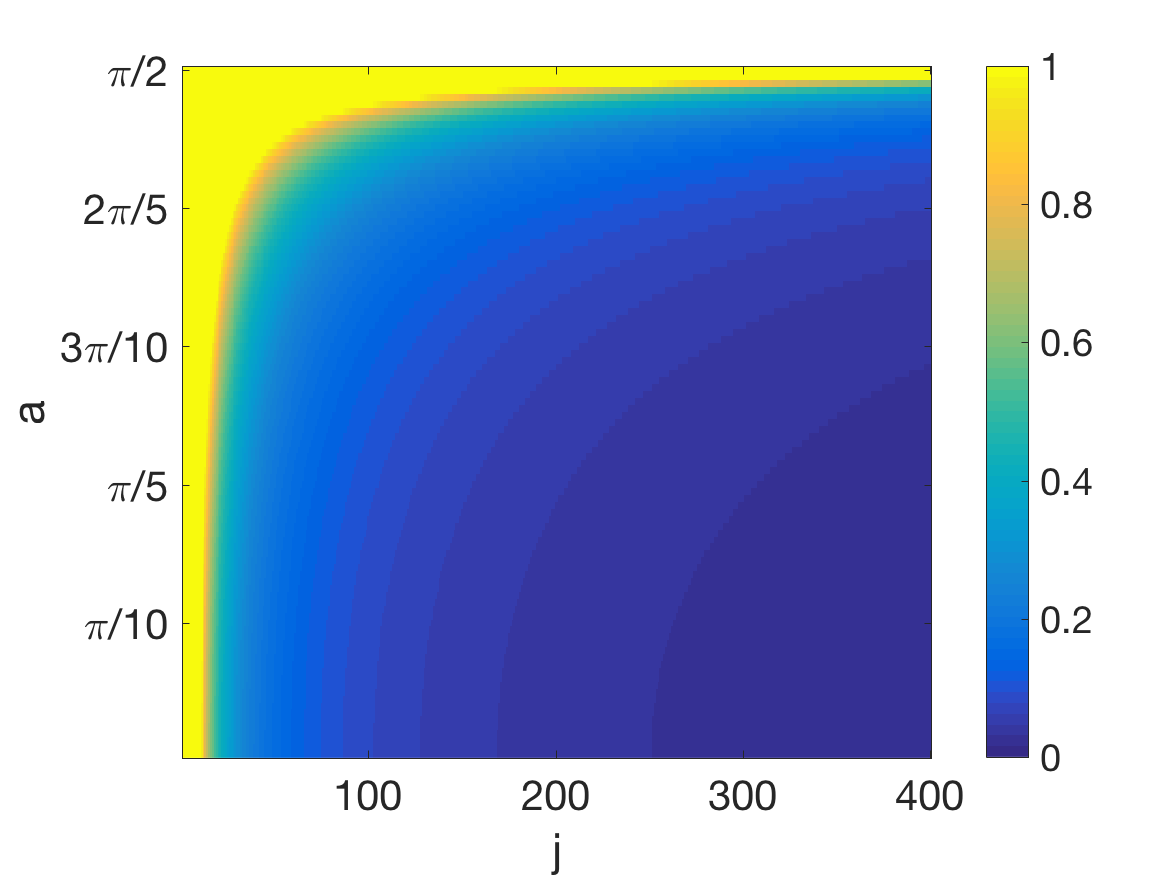}
\includegraphics[width = 0.3\textwidth]{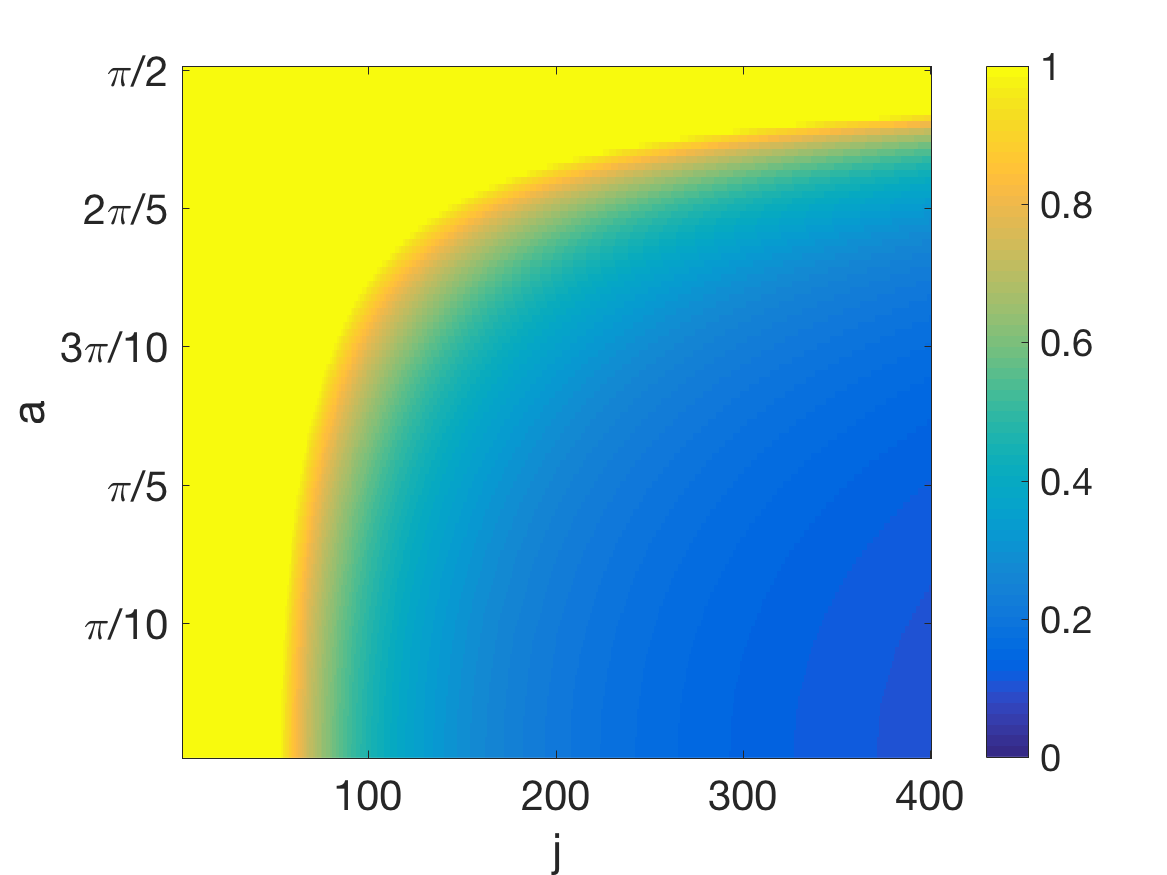}
\includegraphics[width = 0.3\textwidth]{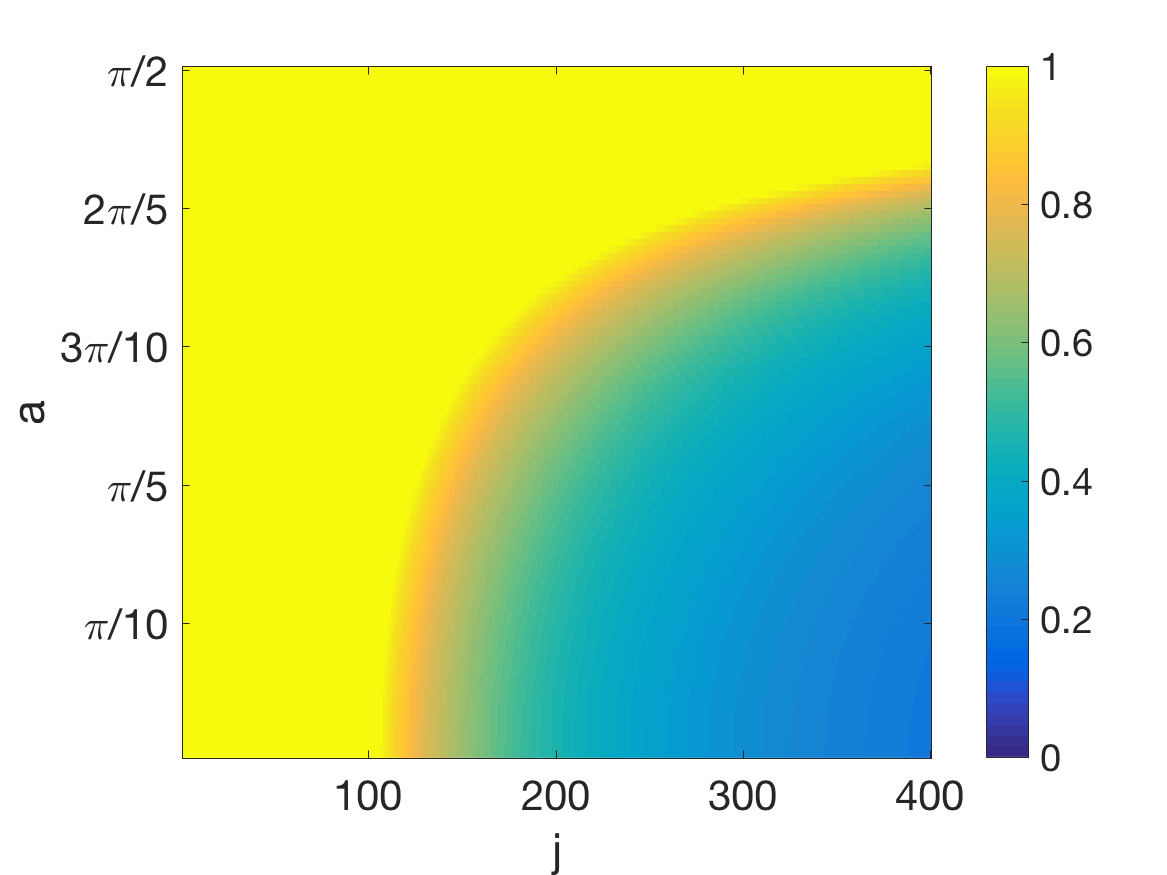}
\caption{For various values of $k_1=k_2$ (the number of hyperplanes intersecting the wedges $G_1$ and $G_2$ respectively), $j$ (the number of hyperplanes separating the wedges $G_1$ and $G_2$), and angles $a$, we plot the bound for $\Pm (\theta \le a)$ given by \Cref{thm:probBound}. From left to right, the plots use $k_1 = 10, 50,$ and $100$ respectively.   }
\label{fig:probBound}
\end{figure}

Ideally, we would like to find a lower bound on the angle $\theta$ between $\tilde \g_1$ and $\tilde \g_2$ that depends on $k_1, k_2$, and $j$. Unfortunately, the explicit form of the resulting angle is relatively complicated. We can simplify the denominator of \Cref{eqn:angle} by using the bounds $||\tilde \g_i||_2\ge j$. We expect this bound to be quite loose, if not trivial, when $j$ is small, but to provide a reasonable bound for larger $j$. With this simplification,
\begin{equation}
\begin{split}
\cos(\theta)\le \frac{\left(j+\sum_{h=1}^{k_1}\frac{u_h(1-u_h)}{(u_h+1)^2}
+ \sum_{h=1}^{k_2}\frac{1-u_h'}{(1+u_h')^2}\right)\left(\sum_{h=1}^{k_1}\frac{u_h(1-u_h)}{(u_h+1)^2}
+ \sum_{h=1}^{k_2}\frac{1-u_h'}{(1+u_h')^2}\right)}{j^2}\\
+\frac{\left(\sum_{h=1}^{k_1}\frac{1-u_h}{(u_h+1)^2}
+ \sum_{h=1}^{k_2}\frac{u_h'(1-u_h')}{(1+u_h')^2}\right)\left(j+\sum_{h=1}^{k_1}\frac{1-u_h}{(u_h+1)^2}
+ \sum_{h=1}^{k_2}\frac{u_h'(1-u_h')}{(1+u_h')^2}\right)}{j^2}.
\label{eqn:angleBound}
\end{split}
\end{equation}
For this simplified bound, taking an expectation is a straightforward calculation. See \Cref{sec:app} for details. We eventually arrive at the bound
\begin{equation}
\begin{split}
\E(\cos(\theta))&\le\frac{(k_1+k_2)(2\log 2-1)}{j}+\frac{(k_1^2+k_2^2)(10(\log 2)^2 -14\log 2+5)}{j^2}\\
&+\frac{4k_1k_2(1-\log 2)(3\log 2-2)+(k_1+k_2)(-2/3+8\log 2-10(\log 2)^2)}{j^2}
\label{eqn:expBound}
\end{split}
\end{equation}
Using Markov's inequality, for $a\in(0,\pi/2)$,
\begin{equation}
\Pm(\theta\le a) = \Pm\left[\cos(\theta)\ge \cos(a)\right] \le \frac{\E(\cos(\theta))}{\cos(a)} .
\label{eqn:markov}
\end{equation}
Although this bound is relatively loose, for sufficiently small $a$ and large $j$, the probability that $\theta\le a$ is small.   We summarize this result in \Cref{thm:probBound}.  More visually appealing, \Cref{fig:probBound} gives the probabilities that result from combining \Cref{eqn:expBound} and \Cref{eqn:markov} for a variety of hyperplane combinations and angles $a$. 

\begin{theorem}
Suppose data is distributed as in \Cref{fig:setup}, where points from classes 1 and 2 are uniformly distributed within the wedges $G_1$ and $G_2$ respectively. Suppose that the angles $A_1$ and $A_2$ are equal. Let $k_1$ and $k_2$ be the number of hyperplanes that intersect $G_1$ and $G_2$ respectively. Let $j$ be the number of hyperplanes that separate $G_1$ and $G_2$.  Consider the points $\x_1$ in class 1 and $\x_2$ in class 2 as shown in \Cref{fig:setup}. The angle $\theta$ between the $\tilde \r$ vectors for $\x_1$ and $\x_2$ after a single iteration of \SCB\ with one level $L$ satisfies the following inequality,
\[\Pm(\theta\le a)  \le \frac{C_1j(k_1+k_2)+C_2(k_1^2+k_2^2)+C_3k_1k_2+C_4(k_1+k_2)}{j^2\cos(a)},\]
where 
\begin{align*}
&C_1 =2(\log 2)-1, &C_2=10(\log 2)^2 -14\log 2+5,\\
&C_3 = 4(1-\log 2)(3\log 2-2), & C_4 = -10(\log 2)^2+8\log 2-2/3.
\end{align*}
\label{thm:probBound}
\end{theorem}

\section{\Cref{algo:Train} for data preprocessing and dimension reduction}\label{sec:prepr}
We remark here briefly about another potential strategy using the output of the \SCB\ approach.  Although this is not the focus of the current work, it may lead to fruitful future directions.  The idea is to use 
 the output from \SCB\ and then apply other established classification methods such as SVM \cite{cortes1995support} to the $\tilde \r$ vectors. Considering SVM specifically, we find that this strategy can perform better than SVM applied directly to the data.

\begin{figure}
\centering
\includegraphics[width=0.47\textwidth]{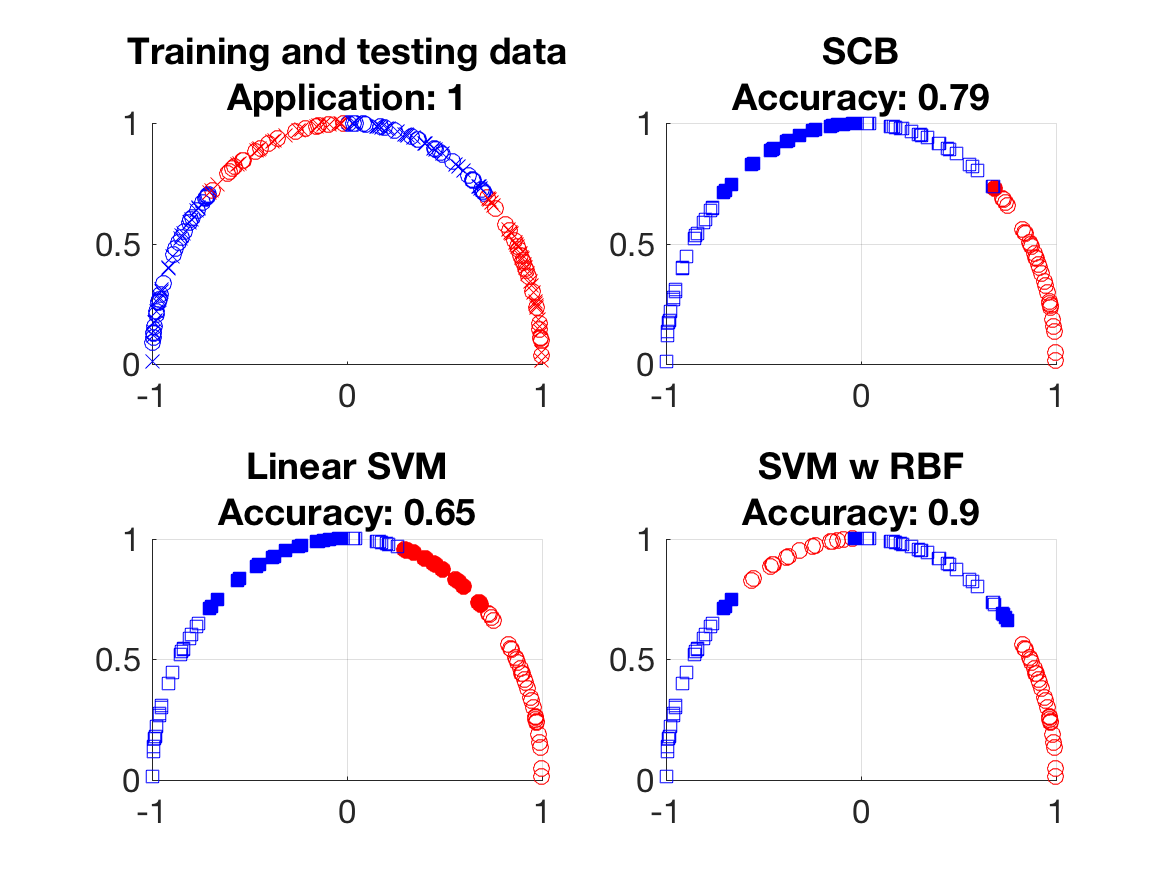}
\includegraphics[width=0.47\textwidth]{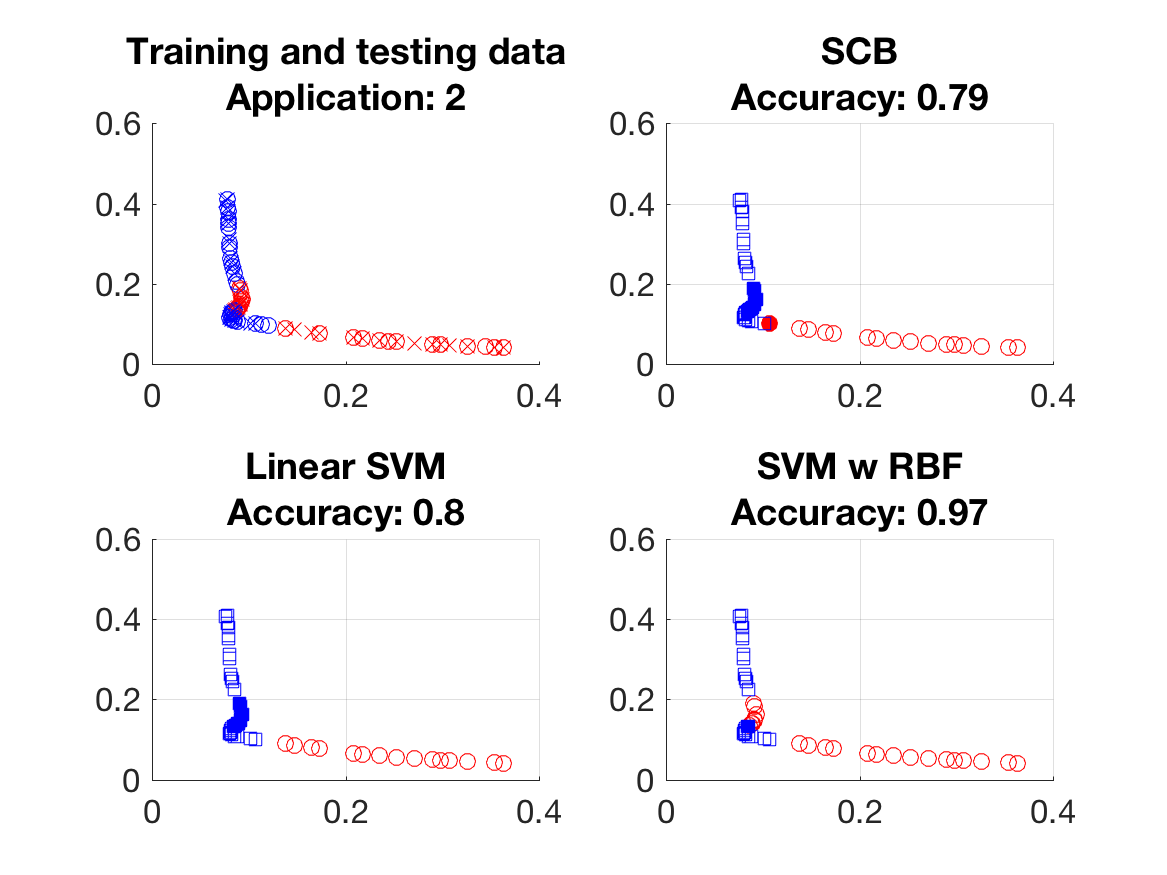}
\caption{The four plots on the left display accuracies and predictions made via various methods for the data given in the upper left-most plot. In the plots of the training and testing data, circles indicate training data and crosses indicate test data. Filled markers indicate that a given method misclassified that particular data point. The methods considered are \SCB\, and SVM with both a linear and RBF kernel. The right set of four plots display accuracies and predictions made via the same set of methods applied to the $\tilde \r$ values from a single application of \SCB\ with a single level ($L=1$) and $m= 100$ measurements.}
\label{fig:SVM_Meta_2D_arc_1layer}
\end{figure}

First, consider a simple example with the synthetic data shown in the upper left plot of \Cref{fig:SVM_Meta_2D_arc_1layer}. Applying SVM with a linear kernel \cite{friedman2001elements,cortes1995support} unsurprisingly performs poorly, achieving an accuracy of 65\%. An RBF kernel \cite{buhmann2003radial,friedman2001elements} SVM performs much better, achieving an accuracy of 90\%. Applying SVM instead to the $\tilde \r_1$ values of the training data produced via \SCB\ with a single level $L$ and $m=100$ measurements leads to 80\% accuracy using a linear kernel and 97\% accuracy using an RBF kernel. Thus, applying SVM to the $\tilde \r_1$ values as opposed to the original data leads to an improvement in accuracy of 15\% for SVM with a linear  and 7\% for SVM with an  RBF kernel. 

\begin{figure}
\centering
\includegraphics[width=0.47\textwidth]{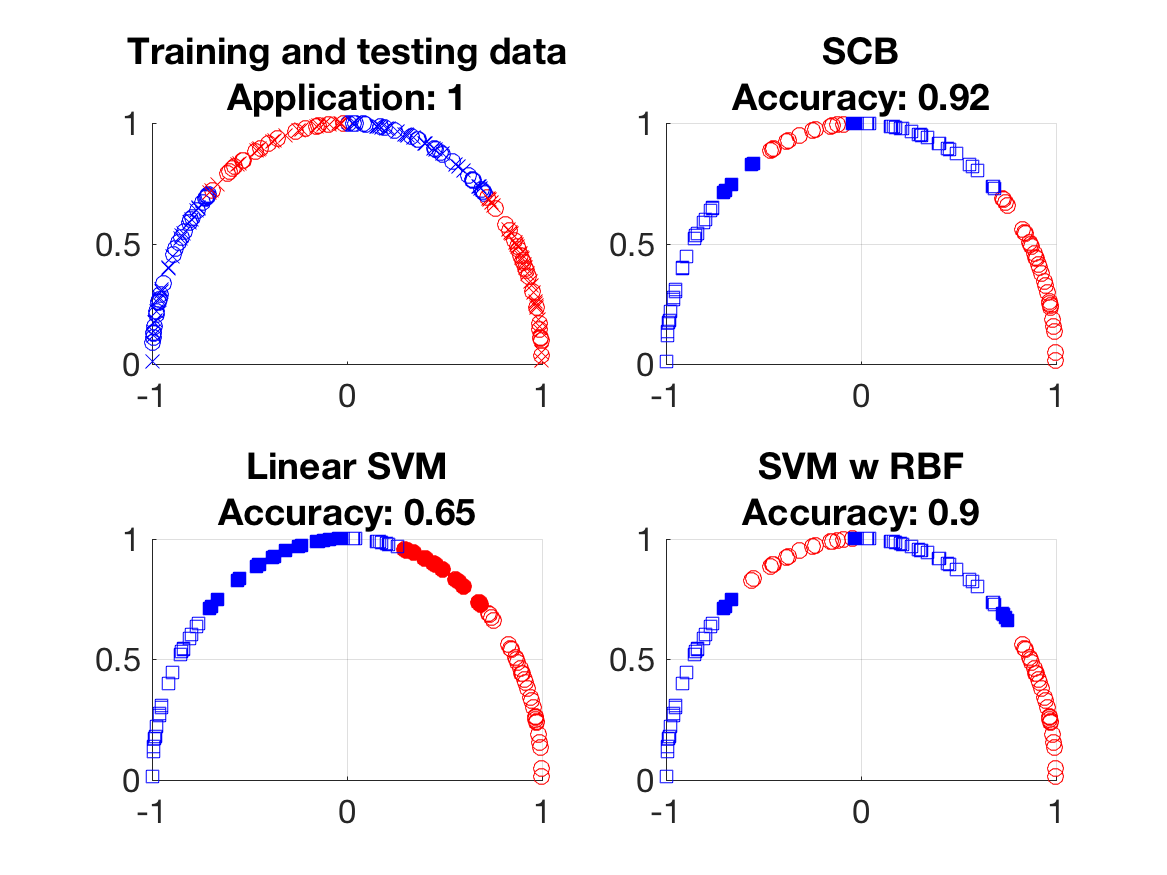}
\includegraphics[width=0.47\textwidth]{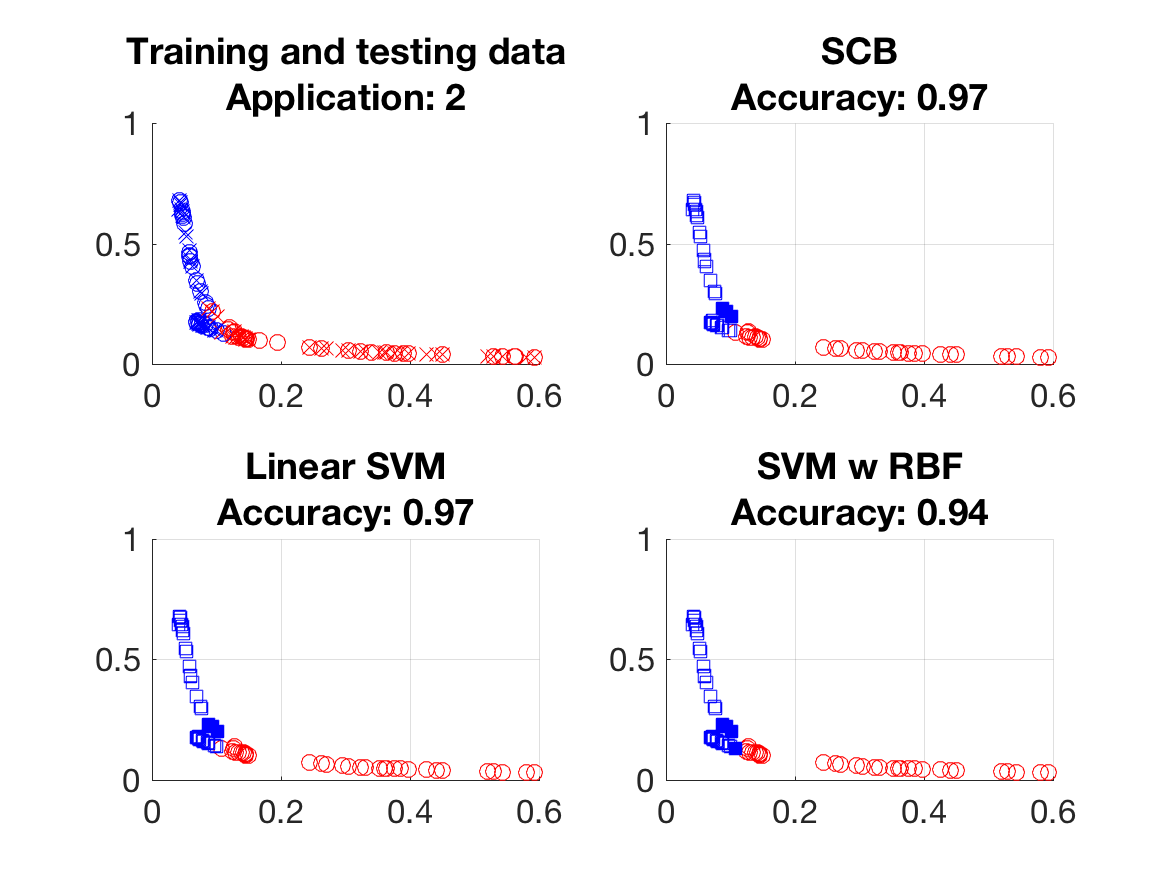}
\caption{The four plots on the left display accuracies and predictions made via various methods for the data given in the upper left-most plot. In the plots of the training and testing data, circles indicate training data and crosses indicate test data. Filled markers indicate that a given method misclassified that particular data point. The methods considered are \SCB\, and SVM with both a linear and RBF kernel. The right set of four plots display accuracies and predictions made via the same set of methods applied to the $\tilde \r$ values from \SCB\ at the first application. $L=4$ levels and $m=200$ measurements are used for each application of \SCB. }
\label{fig:SVM_Meta_2D_arc_4layers}
\end{figure}

For the same initial data, if we increase the number of levels $L$ used in \SCB\ to four and the number of measurements to $m=200$, the accuracies of SVM trained on the resulting $\tilde \r_1$ values are 97\% with a linear kernel and 94\% with an RBF kernel (\Cref{fig:SVM_Meta_2D_arc_4layers}). The respective accuracies are improved by 21\% and 4\% respectively as compared to SVM applied to the original data. This increase in the number of levels $L$ and measurements $m$ also leads to improved performance for both \SCB\ and ISCB with two applications. 
Note that if \SCB\ is able to perfectly classify the training data points, then SVM with a linear kernel trained on the $\tilde \r_1$ values of the training points will also perfectly classify the training data points, as the $\tilde \r_1$ values of the training points will be linearly separable.

•

•

\section{Conclusion}
We have illustrated that iterative applications of SCB of \cite{NSW17Simple} lead to improved classification accuracies as compared to a single application in a variety of settings. Numerical experiments on the MNIST, YaleB, and Norb datasets support this claim. Experiments and theoretical analyses on synthetic data in simple settings demonstrate the effects of multiple iterations on the data and predictions. These examples also highlight simple situations in which the ISCB framework excels. 
We also demonstrate that an application of SCB can be used as a dimension reduction or data preprocessing technique to improve the performance of other classification methods such as SVM.

\section*{Acknowledgments}
The authors would like to thank Tina Woolf for help with the initial code used for the \SCB\ method.
The authors are grateful to and were partially supported by NSF CAREER DMS $\#$1348721 and NSF BIGDATA DMS $\#$1740325. 

\appendix
\section{Detailed calculations for \Cref{sec:Thm1Theory}}
\label{sec:app}
In this section, we provide details for calculating \Cref{eqn:expBound}. Let 
\[K_{11} =\sum_{h=1}^{k_1}\frac{u_h(1-u_h)}{(u_h+1)^2},\; K_{12} = \sum_{h=1}^{k_{1}}\frac{1-u_h}{(u_h+1)^2}  ,\; K_{21} = \sum_{h=1}^{k_2}\frac{1-u_h'}{(1+u_h')^2},\; K_{22} = \sum_{h=1}^{k_2}\frac{u_h'(1-u_h')}{(1+u_h')^2},\]
where $u_h$ and $u_h'$ are i.i.d. uniformly random variables between zero and one.
We can then rewrite \Cref{eqn:angleBound} as 
\begin{multline}
\cos(\theta)\le \frac{(j+K_{11}+K_{21})(K_{11}+K_{21}) + (j+K_{12}+K_{22})(K_{12}+K_{22})}{j^2} \\
=\frac{j(K_{11}+K_{21}+K_{12}+K_{22})+K_{11}^2+2K_{11}K_{21}+K_{21}^2 +K_{12}^2 +2K_{12}K_{22}+K_{22}^2}{j^2}.
\label{eqn:angleBoundK}
\end{multline}

We then require the expectation of each term in the numerator. Since $u_h$ and $u_h'$ are i.i.d., $\E K_{11}K_{21} = \E K_{11}\E K_{21}$.  
Straightforward integral calculations lead to the following expected values:
\begin{align*}
&\E\left(\frac{u_h(1-u_h)}{(u_h+1)^2}\right) = 3\log 2-2 ,
&\E\left(\frac{1-u_h}{(u_h+1)^2}\right)= 1-\log 2, \\
&\E\left(\frac{u_h^2(1-u_h)^2}{(u_h+1)^4}\right) = 25/6 -6\log 2, 
&\E\left(\frac{(1-u_h)^2}{(u_h+1)^4}\right)= 1/6 .
\end{align*}
We then have the following expectations:
\begin{align*}
\E K_{11}&=k_1(3\log 2-2) \\
\E K_{12}&=k_1(1-\log 2)\\
\E K_{11}^2&=k_1(k_1-1)(3\log 2-2)^2 + k_1(25/6 - 6\log 2)\\
\E K_{12}^2&=k_1(k_1-1)(1-\log 2)^2 + k_1(1/6).
\end{align*}
$\E K_{22}, \E K_{21}, \E K_{22}^2$, and $\E K_{21}^2$ take the same forms with $k_2$ replacing $k_1$.

Taking the expectation of \Cref{eqn:angleBoundK},
\begin{align*}
\E(\cos(\theta))&\le\frac{(k_1+k_2)(2\log 2-1)}{j}\\
&+\frac{(k_1^2-k_1+k_2^2-k_2)(3\log 2-2)^2+(k_1^2-k_1+k_2^2-k_2)(1-\log 2)^2}{j^2}\\
&+\frac{4k_1k_2(1-\log 2)(3\log 2-2)+(k_1+k_2)(1/6 + 25/6-6\log 2)}{j^2}\\
&\le \frac{(k_1+k_2)(2\log 2-1)}{j}\\
&+\frac{(k_1^2-k_1+k_2^2-k_2)(10(\log 2)^2 -14\log 2+5)}{j^2}\\
&+\frac{4k_1k_2(1-\log 2)(3\log 2-2)+(k_1+k_2)(13/3-6\log 2)}{j^2}\\
&\le \frac{(k_1+k_2)(2\log 2-1)}{j}+\frac{(k_1^2+k_2^2)(10(\log 2)^2 -14\log 2+5)}{j^2}\\
&+\frac{4k_1k_2(1-\log 2)(3\log 2-2)+(k_1+k_2)(-2/3+8\log 2-10(\log 2)^2)}{j^2},
\end{align*}
providing the desired bound.

\bibliographystyle{plain}
\bibliography{dmbib}

\end{document}